\documentclass[10pt,journal,cspaper,compsoc]{IEEEtran}
\ifCLASSOPTIONcompsoc
  \usepackage[nocompress]{cite}
\else
\fi
\ifCLASSINFOpdf
    \usepackage[pdftex]{graphicx}
    \DeclareGraphicsExtensions{.pdf,.jpeg,.png}
\else
    \usepackage[dvips]{graphicx}
    \DeclareGraphicsExtensions{.eps}
\fi
\usepackage[cmex10]{amsmath}
\usepackage{algorithmic}
\usepackage{url}

\usepackage{multirow}
\usepackage{algorithm}

\usepackage{color}
\usepackage{amsmath}
\usepackage{amssymb}
\usepackage{booktabs}
\usepackage{comment}

\begin{document}
%
\title{Mining Interpretable AOG Representations from Convolutional Networks via Active Question Answering}

%
%

\author{Quanshi Zhang, Ruiming Cao, Ying Nian Wu, and Song-Chun Zhu \textit{Fellow, IEEE}
\IEEEcompsocitemizethanks{\IEEEcompsocthanksitem Quanshi Zhang is with the Shanghai Jiao Tong University, Shanghai, China. Ruiming Cao, Ying Nian Wu, and Song-Chun Zhu are with the University of California, Los Angeles, USA}
}

%
%

\markboth{IEEE TRANSACTIONS ON PATTERN ANALYSIS AND MACHINE INTELLIGENCE, in submission}%
{Shell \MakeLowercase{\textit{et al.}}: Bare Demo of IEEEtran.cls for Computer Society Journals}
%


\IEEEcompsoctitleabstractindextext{%
\begin{abstract}
In this paper, we present a method to mine object-part patterns from conv-layers of a pre-trained convolutional neural network (CNN). The mined object-part patterns are organized by an And-Or graph (AOG). This interpretable AOG representation consists of a four-layer semantic hierarchy, \emph{i.e.} semantic parts, part templates, latent patterns, and neural units. The AOG associates each object part with certain neural units in feature maps of conv-layers. The AOG is constructed in a weakly-supervised manner, \emph{i.e.} very few annotations (\emph{e.g.} 3--20) of object parts are used to guide the learning of AOGs. We develop a question-answering (QA) method that uses active human-computer communications to mine patterns from a pre-trained CNN, in order to incrementally explain more features in conv-layers. During the learning process, our QA method uses the current AOG for part localization. The QA method actively identifies objects, whose feature maps cannot be explained by the AOG. Then, our method asks people to annotate parts on the unexplained objects, and uses answers to discover CNN patterns corresponding to the newly labeled parts. In this way, our method gradually grows new branches and refines existing branches on the AOG to semanticize CNN representations. In experiments, our method exhibited a high learning efficiency. Our method used about $1/6$--$1/3$ of the part annotations for training, but achieved similar or better part-localization performance than fast-RCNN methods.
\end{abstract}


}

\maketitle
\IEEEdisplaynotcompsoctitleabstractindextext
\IEEEpeerreviewmaketitle

\section{Introduction}

Convolutional neural networks~\cite{CNN,CNNImageNet,ResNet,DenseNet} (CNNs) have achieved superior performance in many visual tasks, such as object detection and segmentation. However, in real-world applications, current neural networks still suffer from low interpretability of their middle-layer representations and data-hungry learning methods.


Thus, the objective of this study is to mine thousands of \textit{latent patterns} from the mixed representations in conv-layers. Each latent pattern corresponds to a constituent region or a contextual region of an object part. We use an interpretable graphical model, namely an And-Or graph (AOG), to organize latent patterns hidden in conv-layers. The AOG maps implicit latent patterns to explicit object parts, thereby explaining the hierarchical representation of objects. We use very few (\emph{e.g.} 3--20) part annotations to mine latent patterns and construct the AOG to ensure high learning efficiency.

\begin{figure*}[t]
\centering
\includegraphics[width=\linewidth]{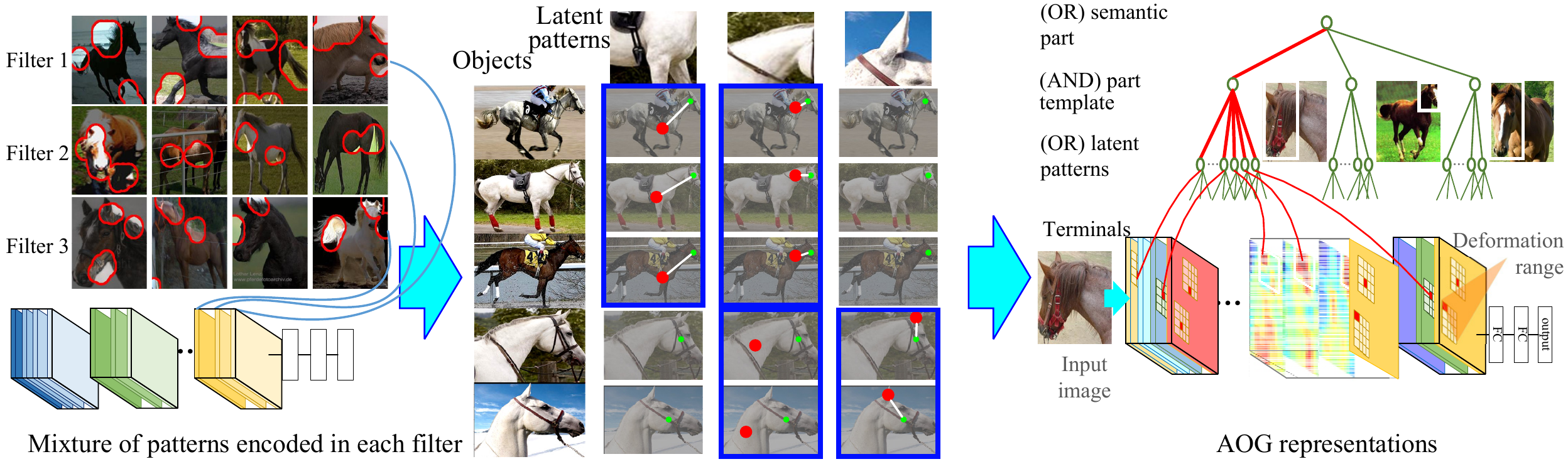}
\caption{Mining part-based AOG representations from CNN representations. (left) Each filter in a conv-layer usually encodes a mixture of patterns, which makes conv-layer representations a black box. The same filter may be activated by different parts on different objects. (middle) We disentangle CNN feature maps and mine latent patterns of object parts. White lines indicate the spatial relationship between a latent pattern's neural activation and the ground-truth position of an object part (head). (right) We grow an AOG on the CNN to associate CNN units with certain semantic parts (the horse head, here). Red lines in the AOG indicate a parse graph that associates certain CNN units with a semantic part.}
\label{fig:rawMapToModel}
\end{figure*}

As shown in Fig.~\ref{fig:rawMapToModel}, compared to ordinary CNN representations where each filter encodes a mixture of textures and parts (evaluated by \cite{Interpretability}), we extract clear object-part representations from CNN features. Our weakly-supervised learning method enables people to model objects or object parts on-the-fly, thereby ensuring broad applicability.

\textbf{And-Or graph representations:}{\verb| |} As shown in Fig.~\ref{fig:rawMapToModel}, the AOG represents a semantic hierarchy on the top of conv-layers, which consists of four layers, \emph{i.e.} the \textit{semantic part}, \textit{part templates}, \textit{latent patterns}, to \textit{CNN units}. In the AOG, AND nodes represent compositional regions of a part, and OR nodes represent a list of alternative template/deformation candidates for a local region.
\begin{itemize}
\item Layer 1: the top \textit{semantic part} node is an OR node, whose children represent template candidates for the part.
\item Layer 2: a \textit{part template} in the second layer describes a certain part appearance with a specific pose, \emph{e.g.} a black sheep head from a side view. A part template is an AND node, which uses its children latent patterns to encode its constituent regions.
\item Layer 3: a \textit{latent pattern} in the third layer represents a constituent region of a part (\emph{e.g.} an eye in the head part) or a contextual region (\emph{e.g.} the neck region \emph{w.r.t.} the head). A latent pattern is an OR node, which naturally corresponds to a group of units within the feature map of a certain CNN filter. The latent pattern selects one of its children \textit{CNN units} as the configuration of the geometric deformation.
\item Layer 4: terminal nodes are \textit{CNN units}, \emph{i.e.} raw activation units on feature maps of a CNN filter.
\end{itemize}
In this hierarchy, the AOG maps implicit latent patterns in raw CNN feature maps to explicit semantic parts. We can use the AOG to localize object parts and their constituent regions for hierarchical object parsing. The AOG is interpretable and can be used for communications with human users.

\begin{figure*}[t]
\centering
\includegraphics[width=0.99\linewidth]{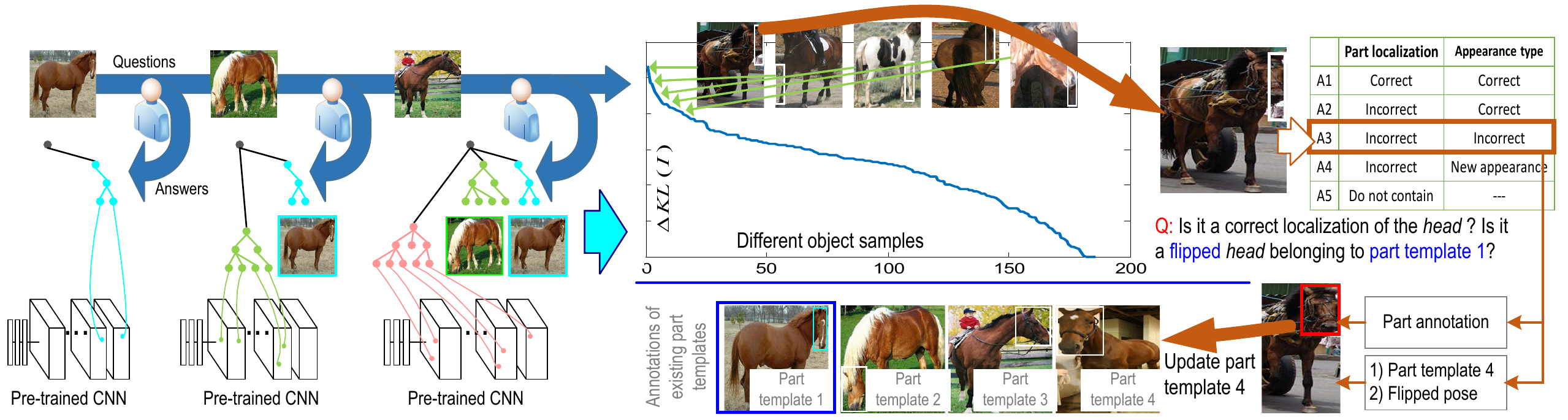}
\caption{Learning an AOG to explain a pre-trained CNN via active question-answering (QA). (left) We mine latent patterns of object parts from the CNN, and organize such patterns into a hierarchical AOG. Our method automatically identifies objects whose parts cannot be well fit current part templates in the AOG, asks about the objects, and uses the answers to mine latent patterns and grow the AOG. (right) Our method sorts and selects objects for QA.}
\label{fig:QA}
\end{figure*}

\textbf{Weakly-supervised learning via active question-answering:}{\verb| |} We propose a new active learning strategy to build an AOG in a weakly-supervised manner. As shown in Fig.~\ref{fig:QA}, we use an active question-answering (QA) process to mine latent patterns from raw feature maps and gradually grow the AOG.

The input is a pre-trained CNN and its training samples (\emph{i.e.} object images without part annotations). The QA method actively discovers the missing patterns in the current AOG and asks human users to label object parts for supervision.

In each step of the QA, we use the current AOG to localize a certain semantic part among all unannotated images. Our method actively identifies object images, which cannot fit well to the AOG. \emph{I.e.} the current AOG cannot explain object parts in these images. Our method estimates the potential gain of asking about each of the unexplained objects, thereby determining an optimal sequence of questions for QA. Note that the QA is implemented based on pre-define ontology, instead of using open-ended questions or answers. As in Fig.~\ref{fig:QA}, the user is asked to provide five types of answers (\emph{e.g.} labeling the correct part position when the AOG cannot accurately localize the part), in order to guide the growth of the AOG. Given each specific answer, our method may either refine the AOG branch of an existing part template or construct a new AOG branch for a new part template.

Based on human answers, we mine latent patterns for new AOG branches as follows. We require the new latent patterns
\begin{itemize}
\item to represent a region highly related to the annotated object parts,
\item to frequently appear in unannotated objects,
\item to consistently keep stable spatial relationships with other latent patterns.
\end{itemize}
Similar requirements were originally proposed in studies of pursuing AOGs, which mined hierarchical object structures from Gabor wavelets on edges~\cite{MiningAOG} and HOG features~\cite{OurICCV15AoG}. We extend such ideas to feature maps of neural networks.

The active QA process mines object-part patterns from the CNN with fewer human supervision. There are three mechanisms to ensure the stability of weakly-supervised learning.
\begin{itemize}
\item Instead of learning all representations from scratch, we transfer patterns in a pre-trained CNN to the target object part, which boosts the learning efficiency. Because the CNN has been trained using numerous images, latent patterns in the AOG are supposed to consistently describe the same part region among different object images, instead of over-fitting to part annotations obtained during the QA process. For example, we use the annotation of a specific tiger head to mine latent patterns. The mined patterns are not over-fitted to the head annotation, but represent generic appearances of different tiger heads. In this way, we can use very few (\emph{e.g.} 1--3) part annotations to extract latent patterns for each part template.
\item It is important to maintain the generality of the pre-trained CNN during the learning procedure. \emph{I.e.} we do not change/fine-tune the original convolutional weights within the CNN, when we grow new AOGs. This allows us to continuously learn new semantic parts from the same CNN, without the model drift.
\item The active QA strategy reduces the excessive usage of the human labor of annotating object parts that have been well explained by the current AOG.
\end{itemize}

In addition, we use object-level annotations for pre-training, considering the following two facts: 1) Only a few datasets~\cite{SemanticPart,CUB200} provide part annotations, and most benchmark datasets~\cite{PascalVOC,ImageNet,MSCOCO} mainly have annotations of object bounding boxes. 2) More crucially, real-world applications may focus on various object parts on-the-fly, and it is impractical to annotate a large number of parts for each specific task.

This paper makes the following three contributions.

\noindent
1) From the perspective of object representations, we semanticize a pre-trained CNN by mining reliable latent patterns from noisy feature maps of the CNN. We design an AOG to represent the semantic hierarchy inside conv-layers, which associates implicit neural patterns with explicit semantic parts.

\noindent
2) From the perspective of learning strategies, based on the clear semantic structure of the AOG, we present an active QA method to learn each part template of the object sequentially, thereby incrementally growing AOG branches on a CNN to enrich part representations in the AOG.

\noindent
3) In experiments, our method exhibits superior performance to other baselines of weakly-supervised part localization. For example, our methods with 11 part annotations outperformed fast-RCNNs with 60 annotations on the Pascal VOC Part dataset.

A preliminary version of this paper appeared in \cite{CNNAoG} and \cite{DeepQA}.

\section{Related work}


\textbf{CNN visualization:}{\verb| |} Visualization of filters in a CNN is a direct way of exploring the pattern hidden inside a neural unit. Lots of visualization methods have been used in the literature.

Gradient-based visualization~\cite{CNNVisualization_1,CNNVisualization_2,CNNVisualization_3} estimates the input image that maximizes the activation score of a neural unit. Dosovitskiy~\emph{et al.}~\cite{FeaVisual} proposed up-convolutional nets to invert feature maps of conv-layers to images. Unlike gradient-based methods, up-convolutional nets cannot mathematically ensure the visualization result reflects actual neural representations. In recent years, \cite{olah2017feature} provided a reliable tool to visualize filters in different conv-layers of a CNN.

Zhou~\emph{et al.}~\cite{CNNSemanticDeep} proposed a method to accurately compute the image-resolution receptive field of neural activations in a feature map. Theoretically, the actual receptive field of a neural activation is smaller than that computed using the filter size. The accurate estimation of the receptive field is crucial to understand a filter's representations.

Unlike network visualization, our mining part representations from conv-layers is another choice to interpret CNN representations.

\textbf{Active network diagnosis:}{\verb| |} Going beyond ``passive'' visualization, some methods ``actively'' diagnose a pre-trained CNN to obtain insight understanding of CNN representations.

\cite{CNNAnalysis_1} explored semantic meanings of convolutional filters. \cite{CNNAnalysis_2} evaluated the transferability of filters in intermediate conv-layers. \cite{CNNAnalysis_3,CNNVisualization_5} computed feature distributions of different categories in the CNN feature space. Methods of \cite{visualCNN_grad,visualCNN_grad_2} propagated gradients of feature maps \emph{w.r.t.} the CNN loss back to the image, in order to estimate the image regions that directly contribute the network output. \cite{trust} proposed a LIME model to extract image regions that are used by a CNN to predict a label (or an attribute).

Network-attack methods~\cite{pixelAttack,CNNInfluence,CNNAnalysis_1} diagnosed network representations by computing adversarial samples for a CNN. In particular, influence functions~\cite{CNNInfluence} were proposed to compute adversarial samples, provide plausible ways to create training samples to attack the learning of CNNs, fix the training set, and further debug representations of a CNN. \cite{banditUnknown} discovered knowledge blind spots (unknown patterns) of a pre-trained CNN in a weakly-supervised manner.

Zhang~\emph{et al.}~\cite{CNNBias} developed a method to examine representations of conv-layers and automatically discover potential, biased representations of a CNN due to the dataset bias. Furthermore, \cite{wu2007compositional,yang2009evaluating,wu2011numerical} mined the local, bottom-up, and top-down information components in a model for prediction.

\textbf{CNN semanticization:}{\verb| |} Compared to the diagnosis of CNN representations, semanticization of CNN representations is closer to the spirit of building interpretable representations.

Hu~\emph{et al.}~\cite{LogicRuleNetwork} designed logic rules for network outputs, and used these rules to regularize neural networks and learn meaningful representations. However, this study has not obtained semantic representations in intermediate layers. Some studies extracted neural units with certain semantics from CNNs for different applications. Given feature maps of conv-layers, Zhou~\emph{et al.}~\cite{CNNSemanticDeep,CNNSemanticDeep2} extracted scene semantics. Simon~\emph{et al.} mined objects from feature maps of conv-layers~\cite{ObjectDiscoveryCNN_2}, and learned explicit object parts~\cite{CNNSemanticPart}.

Unlike above research, we aim to explore the entire semantic hierarchy hidden inside conv-layers of a CNN. Because the AOG structure~\cite{MumfordAOG,MiningAOG} is suitable for representing the semantic hierarchy of objects, our method uses an AOG to represent the CNN. In our study, we use semantic-level QA to incrementally mine object parts from the CNN and grow the AOG. Such a ``white-box'' representation of the CNN also guided further active QA. With clear semantic structures, the AOG makes it easier to transfer CNN patterns to other part-based tasks.

\textbf{Unsupervised/active learning:}{\verb| |} Many methods have been developed to learn object models in an unsupervised or weakly supervised manner. Methods of \cite{Gpt_WeaklyCNN,WeaklyMIL,OurICCV15AoG,ObjectDiscoveryCNN_2} learned with image-level annotations without labeling object bounding boxes. \cite{UnsuperCNN,ChoDiscovery} did not require any annotations during the learning process. \cite{OnlineMetric} collected training data online from videos to incrementally learn models. \cite{Language2VideoAlign,Language2ActionAlign} discovered objects and identified actions from language Instructions and videos. Inspired by active learning~\cite{Active4,i13,Active2}, the idea of learning from question-answering has been used to learn object models~\cite{KB_Fei_Annotation,KB_Fei_InteractionLabel,TuQA}. Branson~\emph{et al.}~\cite{ActivePart} used human-computer interactions to label object parts to learn part models. Instead of directly building new models from active QA, our method uses the QA to mine AOG part representations from CNN representations.

\textbf{AOG for knowledge transfer:} Transferring hidden patterns in the CNN to other tasks is important for neural networks. Typical research includes end-to-end fine-tuning and transferring CNN representations between different categories~\cite{CNNAnalysis_2,CNNSemantic} or datasets~\cite{UnsuperTransferCNN}. In contrast, we believe that a good explanation and transparent representation of parts will create a new possibility of transferring part features. As in \cite{AllenAoG,MiningAOG}, the AOG is suitable to represent the semantic hierarchy, which enables semantic-level interactions between human and neural networks.

\textbf{Modeling ``objects'' vs. modeling ``\textbf{parts}'' in un-/weakly-supervised learning:}{\verb| |} Generally speaking, in the scenario of un-/weakly-supervised learning, it is usually more difficult to model object parts than to represent entire objects. For example, object discovery~\cite{ObjectDiscoveryCNN_1,ObjectDiscoveryCNN_2,ObjectDiscoveryCNN_3} and co-segmentation~\cite{InteractiveCoseg} only require image-level labels without object bounding boxes. Object discovery is mainly implemented by identifying common foreground patterns from the noisy background. People usually consider closed boundaries and common object structure as a strong prior for object discovery.

In contrast to objects, it is difficult to mine true part parsing of objects without sufficient supervision. Up to now, there is no reliable solution to distinguishing semantically meaningful parts from other potential divisions of object parts in an unsupervised manner. In particular, some parts (\emph{e.g.} the abdomen) do not have shape boundaries to determine their shape extent.


\textbf{Part localization/detection vs. semanticizing CNN patterns:} There are two key points to differentiate our study from conventional part-detection approaches. First, most detection methods deal with classification problems, but inspired by graph mining~\cite{OurICCV15AoG,OurSAPPAMI,OurCVPR14Graph}, we mainly focus on a mining problem. \emph{I.e.} we aim to discover meaningful latent patterns to clarify CNN representations. Second, instead of summarizing common knowledge from massive annotations, our method requires very limited supervision to mine latent patterns.


\section{Method}

The overall objective is to sequentially minimize the following three loss terms.
\begin{equation}
{Loss}={Loss}^{\textrm{CNN}}+{Loss}^{\textrm{QA}}+{Loss}^{\textrm{AOG}}
\label{eqn:obj}
\end{equation}
${Loss}^{\textrm{CNN}}$ denotes the classification loss of the CNN.

${Loss}^{\textrm{QA}}$ is referred as to the loss for active QA. Given the current AOG, we use ${Loss}^{\textrm{QA}}$ to actively determine a sequence of questions about objects that cannot be explained by the current AOG, and require people to annotate bounding boxes of new object parts for supervision.

${Loss}^{\textrm{AOG}}$ is designed to learn an AOG for the CNN. ${Loss}^{\textrm{AOG}}$ penalizes 1) the incompatibility between the AOG and CNN feature maps of unannotated objects and 2) part-location errors \emph{w.r.t.} the annotated ground-truth part locations.

It is essential to determine the optimization sequence for the three losses in the above equation. We propose to first learn the CNN by minimizing ${Loss}^{\textrm{CNN}}$ and then build an AOG based on the learned CNN. We use the active QA to obtain new part annotations and use new part annotations to grow the AOG by optimizing ${Loss}^{\textrm{QA}}$ and ${Loss}^{\textrm{AOG}}$ alternatively.

We introduce details of the three losses in the following subsections.

\subsection{Learning convolutional neural networks}

To simplify the story, in this research, we just consider a CNN for single-category classification, \emph{i.e.} identifying object images of a specific category from random images. We use the log logistic loss to learn the CNN.
\begin{equation}
{Loss}^{\textrm{CNN}}=\mathbb{E}_{I\in{\bf I}}\big[{Loss}(\hat{y}_{I},y^{*}_{I})\big]
\end{equation}
where $\hat{y}_{I}$ and $y^{*}_{I}$ denote the predicted and ground-truth labels of an image $I$. If the image $I$ belongs to the target category, then $y^{*}_{I}=+1$; otherwise $y^{*}_{I}=-1$.

\subsection{Learning And-Or graphs}

We are given a pre-trained CNN and its training images without part annotations. We use an active QA process to obtain a small number of annotations of object-part bounding boxes, which will be introduced in Section~\ref{sec:QA}. Based on these inputs, in this subsection, we focus on the approach for learning an AOG to represent the object part.

\subsubsection{And-Or graph representations}

Before the introduction of learning AOGs, we first briefly overview the structure of the AOG and the part parsing (inference) based on the AOG.

As shown in Fig.~\ref{fig:rawMapToModel}, an AOG represents the semantic structure of a part at four layers.
\begin{center}
\begin{tabular}{c|lc}
\hline
Layer \!\!&\!\! Name \!\!&\!\! Node type\\
\hline
1 \!\!&\!\! semantic part \!\!&\!\! OR node\\
2 \!\!&\!\! part template \!\!&\!\! AND node\\
3 \!\!&\!\! latent pattern \!\!&\!\! OR node\\
4 \!\!&\!\! neural unit \!\!&\!\! Terminal node\\
\hline
\end{tabular}
\end{center}
In the AOG, each OR node encodes a list of alternative appearance (or deformation) candidates as children. Each AND node uses its children to represent its constituent regions.

More specifically, the top node is an OR node, which represents a certain semantic part, \emph{e.g.} the head or the tail. The semantic part node encodes some part templates as children. Each part template corresponds to a specific part appearance from a certain perspective. During the inference process, the semantic part (an OR node) selects the best part template among all template candidates to represent the object.

The part template in the second layer is an AND node, which uses its children latent patterns to represent a constituent region or a contextual region \emph{w.r.t.} the part template. The part template encodes spatial relationships between its children.

The latent pattern in the third layer is an OR node, whose receptive field is a square block within the feature map of a specific convolutional filter. The latent pattern takes neural units inside its receptive field as children. Because the latent pattern may appear at different locations in the feature map, the latent pattern uses these neural units to represent its deformation candidates. During the inference process, the latent pattern selects the strongest activated child unit as its deformation configuration.

Given an image $I$\footnote{Because the CNN has demonstrated its superior performance in object detection, we assume that the target object can be well detected by the pre-trained CNN. As in \cite{SemanticPart}, we regard object detection and part localization as two separate processes for evaluation. Thus, to simplify the learning scenario, we crop $I$ only to contain the object, resize it to the image size for CNN inputs, and just focus on the part localization task to simplify the scenario of learning for part localization.}, we use the CNN to compute feature maps of all conv-layers on image $I$. Then, we can use the AOG for hierarchical part parsing. \emph{I.e.} we use the AOG to semanticize the feature maps and localize the target part and its constituent regions in different layers.

The parsing result is illustrated as red lines in Fig.~\ref{fig:rawMapToModel}. From a top-down perspective, the parsing procedure 1) identifies a part template for the semantic part; 2) parses an image region for the selected part template; 3) for each latent pattern under the part template, it selects a neural unit within a specific deformation range to represent this pattern.

\textbf{OR nodes:} Both the top semantic-part node and latent-pattern nodes in the third layer are OR nodes. The parsing process assigns each OR node $u$ with an image region $\Lambda_{u}$ and an inference score $S_{u}$. $S_{u}$ measures the fitness between the parsed region $\Lambda_{u}$ and the sub-AOG under $u$. The computation of $\Lambda_{u}$ and $S_{u}$ for all OR nodes shares the same paradigm.
\begin{equation}
S_{u}=\max_{v\in Child(u)}S_{v},\qquad\Lambda_{u}=\Lambda_{\hat{v}}
\end{equation}
where let $u$ have $m$ children nodes $Child(u)=\{v_{1},v_{2},\ldots,v_{m}\}$. $S_{v}$ denotes the inference score of the child $v$, and $\Lambda_{v}$ is referred to as the image region assigned to $v$. The OR node selects the child with the highest score $\hat{v}={\arg\!\max}_{v\in Child(u)}S_{v}$ as the true parsing configuration. Node $\hat{v}$ propagates its image region to the parent $u$.

More specifically, we introduce detailed settings for different OR nodes.
\begin{itemize}
\item The OR node of the top semantic part contains a list of alternative part templates. We use $top$ to denote the top node of the semantic part. The semantic part chooses a part template to describe each input image $I$.
\item The OR node of each latent pattern $u$ in the third layer naturally corresponds to a square deformation range within the feature map of a convolutional filter of a conv-layer. All neural units within the square are used as deformation candidates of the latent pattern. For simplification, we set a constant deformation range (with a center $\overline{{\bf p}}_{u}$ and a scale of $\frac{h}{3}\times\frac{w}{3}$ in the feature map where $h$ and $w$ ($h=w$) denote the height and width of the feature map) for each latent pattern. $\overline{{\bf p}}_{u}$ is a parameter that needs to be learned. Deformation ranges of different patterns in the same feature map may overlap. Given parsing configurations of children neural units as input, the latent pattern selects the child with the highest inference score as the true deformation configuration.
\end{itemize}

\textbf{AND nodes:} Each part template is an AND node, which uses its children (latent patterns) to represent its constituent or contextual regions. We use $v$ and $Child(v)=\{u_{1},u_{2},\ldots,u_{m}\}$ to denote the part template and its children latent patterns. We learn the average displacement from $\Lambda_{u}$ to $\Lambda_{v}$ among different image, denoted by $\Delta{\bf p}_{u}$, as a parameter of the AOG. Given parsing results of children latent patterns, we use the image region of each child node $\Lambda_{u}$ to infer the region for the parent $v$ based on its spatial relationships. Just like a deformable part model, the parsing of $v$ can be given as
\begin{equation}
S_{v}\!=\!\!\!\!\!\!\!\sum_{u\in Child(v)}\!\!\!\!\!\!\!\big[S_{u}\!+\!S^{\textrm{inf}}(\Lambda_{u}|\Lambda_{v})\big],\;\;\Lambda_{v}\!=\!f(\Lambda_{u_{1}},\ldots,\Lambda_{u_{m}})\!
\end{equation}
where we use parsing results of children nodes to infer the parent part template $v$. $S^{\textrm{inf}}(\Lambda_{u}|\Lambda_{v})$ denotes the spatial compatibility between $\Lambda_{u}$ and $\Lambda_{v}$ \emph{w.r.t.} their average displacement $\Delta{\bf p}_{u}$. Please see the appendix for details of $S^{\textrm{inf}}(\Lambda_{u}|\Lambda_{v})$.

For the region parsing of the part template $v$, we need to estimate two terms, \emph{i.e.} the center position ${\bf p}_{v}$ and the scale $scale_{v}$ of $\Lambda_{v}$. We learn a fixed scale for each part template, which will be introduced in Section~\ref{sec:learnAOG}. In this way, we can simply implement region parsing by computing the region position that maximizes the inference score ${\bf p}_{v}=f(\Lambda_{u_{1}},\Lambda_{u_{2}},\ldots,\Lambda_{u_{m}})={\arg\!\max}_{{\bf p}_{v}}S_{v}$.

\textbf{Terminal nodes (neural units):} Each terminal node under a latent pattern represents a deformation candidate of the latent pattern. The terminal node has a fixed image region, \emph{i.e.} we propagate the neural unit's receptive field back to the image plane as its image region. We compute a neural unit's inference score based on both its neural response value and its displacement \emph{w.r.t.} its parent latent pattern. Please see the appendix for details.

Based on the above node definitions, we can use the AOG to parse each given image $I$ by dynamic programming in a bottom-up manner.

\subsubsection{Learning And-Or graphs}
\label{sec:learnAOG}

The core of learning AOGs is to distinguish reliable latent patterns from noisy neural responses in conv-layers and select reliable latent patterns to construct the AOG.

\textbf{Training data:}{\verb| |} Let ${\bf I}^{\textrm{obj}}\subset{\bf I}$ denote the set of object images of a target category. During the active question-answering, we obtain bounding boxes of the target object part in a small number of images, ${\bf I}^{\textrm{ant}}\!=\!\{I_1,I_2,\ldots,I_{M}\}\subset{\bf I}^{\textrm{obj}}$ among all objects. The other images without part annotations are denoted by ${\bf I}^{\textrm{unant}}={\bf I}^{\textrm{obj}}\setminus{\bf I}^{\textrm{ant}}$. In addition, the question-answering process collects a number of part templates. Thus, for each image $I\in{\bf I}^{\textrm{ant}}$, we annotate $(\Lambda_{top}^{*},v^{*})$, where $\Lambda_{top}^{*}$ denotes the ground-truth bounding box of the part in $I$, and $v^{*}\in Child(top)$ specifies the ground-truth template for the part.

\textbf{Which AOG parameters to learn:}{\verb| |} We can use human annotations to define the first two layers of the AOG. If human annotators specify a total of $m$ different part templates during the annotation process, correspondingly, we can directly connect the top node with $m$ part templates as children. For each part template $v\in Child(top)$, we fix a constant scale for its region $\Lambda_{v}$. \emph{I.e.} if there are $n$ ground-truth part boxes that are labeled for $v$, we compute the average scale among the $n$ part boxes as the constant scale $scale_{v}$.

Thus, the key to AOG construction is to mine children latent patterns for each part template $v$. We need to mine latent patterns from a total of $K$ conv-layers. We select $n_{k}$ latent patterns from the $k$-th ($k=1,2,\ldots,K$) conv-layer, where $K$ and $\{n_{k}\}$ are hyper-parameters. Let each latent pattern $u$ in the $k$-th conv-layer correspond to a square deformation range, which is located in the $D_{u}$-th slice of the conv-layer's feature map. $\overline{\bf p}_{u}$ denotes the center of the range. As analyzed in the appendix, we only need to estimate the parameters of $D_{u},\overline{\bf p}_{u}$ for $u$.

\textbf{How to learn:}{\verb| |} Just like the pattern pursuing in Fig.~\ref{fig:rawMapToModel}, we mine the latent patterns by estimating their best locations $D_{u},\overline{\bf p}_{u}\in{\boldsymbol\theta}$ that maximize the following objective function, where ${\boldsymbol\theta}$ denotes the parameter set of the AOG.
\begin{equation}
\begin{split}
{Loss}^{\textrm{AOG}}=\mathbb{E}_{I\in{\bf I}^{\textrm{ant}}}\big[-S_{top}+L(\Lambda_{top},\Lambda_{top}^{*})\big]\qquad\\
+\lambda^{\textrm{unant}}\mathbb{E}_{I\in{\bf I}^{\textrm{obj}}}\big[-S^{\textrm{unant}}_{\textrm{AOG}}+L^{\textrm{unant}}({\boldsymbol\Lambda}_{\textrm{AOG}})\big]
\end{split}
\label{eqn:LossAOG}
\end{equation}
First, let us focus on the first half of the equation, which learns from part annotations. $S_{top}$ and $L(\Lambda_{top},\Lambda_{top}^{*})$ denote the final inference score of the AOG on image $I$ and the loss of part localization, respectively. Given annotations $(\Lambda_{top}^{*},v^{*})$ on $I$, we get
\begin{equation}
\begin{split}
&S_{top}=\max_{v\in Child(top)}S_{v}\approx S_{v^{*}}\\
&L(\Lambda_{top},\Lambda_{top}^{*})=-\lambda_{v^{*}}\Vert{\bf p}_{top}-{\bf p}^{*}_{top}\Vert
\end{split}
\end{equation}
where we approximate the ground-truth part template $v^{*}$ as the selected part template. We ignore the small probability of the AOG assigning an annotated image with an incorrect part template to simplify the computation. The part-localization loss $L(\Lambda_{top},\Lambda_{top}^{*})$ measures the localization error between the parsed part region ${\bf p}_{top}$ and the ground truth ${\bf p}^{*}_{top}={\bf p}(\Lambda_{top}^{*})$.

The second half of Equation~(\ref{eqn:LossAOG}) learns from objects without part annotations.
\begin{equation}
\begin{split}
S^{\textrm{unant}}_{\textrm{AOG}}&={\sum}_{u\in Child(v^{*})}S^{\textrm{unant}}_{u}\\
L^{\textrm{unant}}({\boldsymbol\Lambda}_{\textrm{AOG}})&={\sum}_{u\in Child(v^{*})}\lambda^{\textrm{close}}\Vert\Delta{\bf p}_{u}\Vert^2
\end{split}
\label{sec:unsuper}
\end{equation}
where the first term $S^{\textrm{unant}}_{\textrm{AOG}}$ denotes the inference score at the level of latent patterns without ground-truth annotations of object parts. Please see the appendix for the computation of $S^{\textrm{unant}}_{u}$. The second term $L^{\textrm{unant}}({\boldsymbol\Lambda}_{\textrm{AOG}})$ penalizes latent patterns that are far from their parent $v^{*}$. This loss encourages the assigned neural unit to be close to its parent latent pattern. We assume that 1) latent patterns that frequently appear among unannotated objects may potentially represent stable part appearance and should have higher priorities; and that 2) latent patterns spatially closer to their parent part templates are usually more reliable.

When we set $\lambda_{v^{*}}$ to a constant $\lambda^{\textrm{inf}}\sum_{k=1}^{K}n_{k}$, we can transform the learning objective in Equation~(\ref{eqn:LossAOG}) as follows.
\begin{equation}
\forall v\in Child(top), \quad\max_{{\boldsymbol\theta}_{v}}{\bf L}_{v},\quad {\bf L}_{v}\!=\!\!\!\!\!\!\!\!\sum_{u\in Child(v)}\!\!\!\!\!\!\!Score(u)
\label{eqn:subAOG}
\end{equation}
where {$Score(u)\!=\!\mathbb{E}_{I\in{\bf I}_{v}}[S_{u}+S^{\textrm{inf}}(\Lambda_{u}|\Lambda^{*}_{v})]$
$+\mathbb{E}_{I'\in{\bf I}^{\textrm{obj}}}$ $\lambda^{\textrm{unant}}[S^{\textrm{unant}}_{u}-\lambda^{\textrm{close}}\Vert\Delta{\bf p}_{u}\Vert^2]$}. ${\boldsymbol\theta}_{v}\subset{\boldsymbol\theta}$ denotes the parameters for the sub-AOG of the part template $v$. We use ${\bf I}_{v}\subset{\bf I}^{\textrm{ant}}$ to denote the subset of images that are annotated with $v$ as the ground-truth part template.

\textbf{Learning the sub-AOG for each part template:}{\verb| |} Based on Equation~(\ref{eqn:subAOG}), we can mine the sub-AOG for each part template $v$, which uses this template's annotations on images $I\in{\bf I}_{v}\subset{\bf I}^{\textrm{ant}}$, as follows.

\noindent
1) We first enumerate all possible latent patterns corresponding to the $k$-th CNN conv-layer ($k=1,\ldots,K$), by sampling all pattern locations \emph{w.r.t.} $D_{u}$ and $\overline{\bf p}_{u}$.

\noindent
2) Then, we sequentially compute $\Lambda_{u}$ and $Score(u)$ for each latent pattern.

\noindent
3) Finally, we sequentially select a total of $n_{k}$ latent patterns. In each step, we select $\hat{u}\!=\!{\arg\!\max}_{u\in Child(v)}\Delta{\bf L}_{v}$. \emph{I.e.} we select latent patterns with top-ranked values of {$Score(u)$} as children of part template $v$.

\subsection{Learning via active question-answering}
\label{sec:QA}

We propose a new learning strategy, \emph{i.e.} active QA, which is more efficient than conventional batch learning. The QA-based learning algorithm actively detects blind spots in feature representations of the model and ask questions for supervision. In general, blind spots in the AOG include 1) neural-activation patterns in the CNN that have not been encoded in the AOG and 2) inaccurate latent patterns in the AOG. The unmodeled neural patterns potentially reflect new part templates, while inaccurate latent patterns correspond to sub-optimized part templates.

As an interpretable representation of object parts, the AOG can represent blind spots using linguistic description. We design five types of answers to project these blind spots onto semantic details of objects. Our method selects and asks a series of questions. We then collect answers from human users, in order to incrementally grow new AOG branches to explain new part templates and refine existing AOG branches of part templates.

Our approach repeats the following QA process. As shown in Fig.~\ref{fig:QA}, at first, we use the current AOG to localize object parts on all unannotated objects of a category. Based on localization results, the algorithm selects and asks about the object $I$, from which the AOG can obtain the most information gain. A question {$q\!=\!(I,\hat{v},\Lambda_{\hat{v}})$} requires people to determine whether our approach predicts the correct part template $\hat{v}$ and parses a correct region $\Lambda_{top}=\Lambda_{\hat{v}}$ for the part. Our method expects one of the following answers.

\textbf{Answer 1:} the part detection is correct. \textbf{Answer 2:} the current AOG predicts the correct part template in the parse graph, but it does not accurately localize the part. \textbf{Answer 3:} neither the part template nor the part location is correctly estimated. \textbf{Answer 4:} the part belongs to a new part template. \textbf{Answer 5:} the target part does not appear in the image. In particular, in case of receiving Answers~2--4, our method will ask people to annotate the target part. In case of getting Answer 3, our method will require people to specify its part template and whether the object is flipped. Our method uses new part annotations to refine (for Answers 2--3) or create (for Answer 4) an AOG branch of the annotated part template based on Equation~(\ref{eqn:LossAOG}).

\subsubsection{Question ranking}

The core of the QA-based learning is to select a sequence of questions that reduce the uncertainty of part localization the most. Therefore, in this section, we design a loss function to measure the incompatibility between the AOG and real part appearances in object samples. Our approach predicts the potential gain (decrease of the loss) of asking about each object. Objects with large gains usually correspond to not well explained CNN neural activations. Note that annotating a part in an object may also help localize parts on other objects, thereby leading to a large gain. Thus, we use a greedy strategy to select a sequence of questions {$\Omega=\{q_{i}|i=1,2,\ldots\}$}, \emph{i.e.} asking about the object that produces the most gain in each step.

For each object image $I$, we use {${\bf P}(y|I)$} and {${\bf Q}(y|I)$} to denote the prior distribution and the estimated distribution of an object part on $I$, respectively. A label {$y\in\{+1,-1\}$} indicates whether $I$ contains the target part. The AOG estimates the probability of object $I$ containing the target part as {${\bf Q}(y\!=\!+1|I)\!=\!\frac{1}{Z}\exp[\beta S_{top}]$}, where $Z$ and $\beta$ are parameters for scaling (see Section~\ref{sec:implement} for details); {${\bf Q}(y=-1|I)\!=\!1-{\bf Q}(y=+1|I)$}. Let {${\bf I}^{\textrm{ant}}$} denote the set of objects without being asked during previous QA. For each asked object {$I\in{\bf I}^{\textrm{ant}}$}, we set its prior distribution {${\bf P}(y=+1|I)=1$} if $I$ contains the target part; {${\bf P}(y=+1|I)=0$} otherwise. For each un-asked object {$I\in{\bf I}^{\textrm{unant}}$}, we set its prior distribution based on statistics of previous answers, {${\bf P}(y=+1|I)=\mathbb{E}_{I'\in{\bf I}^{\textrm{ant}}}{\bf P}(y=+1|I')$}. Therefore, we formulate the loss function as the KL divergence between the prior distribution {${\bf P}$} and the estimated distribution {${\bf Q}$}.
\begin{equation}
\begin{split}
\!\!{Loss}^{\textrm{QA}}\!\!\!=\!{\bf KL}({\bf P}\Vert{\bf Q})\!=\!&\sum_{I\in{\bf I}^{\textrm{obj}}}\sum_{y}{\bf P}(y,I)\log\frac{{\bf P}(y,I)}{{\bf Q}(y,I)}\!\!\!\!\!\!\!\\
=&\lambda\sum_{I\in{\bf I}^{\textrm{obj}}}\sum_{y}{\bf P}(y|I)\log\frac{{\bf P}(y|I)}{{\bf Q}(y|I)}
\end{split}
\end{equation}
where {${\bf P}(y,I)\!=\!{\bf P}(y|I)P(I)$; ${\bf Q}(y,I)\!=\!{\bf Q}(y|I)P(I)$; $\lambda=P(I)\!=\!1/\vert{\bf I}^{\textrm{obj}}\vert$} is a constant prior probability for $I$.

We keep modifying both the prior distribution ${\bf P}$ and the estimated distribution ${\bf Q}$ during the QA process. Let the algorithm select an unannotated object {$\tilde{I}\in{\bf I}^{\textrm{unant}}={\bf I}^{\textrm{obj}}\setminus{\bf I}^{\textrm{ant}}$} and ask people to label its part. The annotation would encode part representations of $\tilde{I}$ into the AOG and significantly change the estimated distribution for objects that are similar to $\tilde{I}$. For each object $I'\in{\bf I}^{\textrm{obj}}$, we predict its estimated distribution after a new part annotation as
\begin{equation}
\begin{split}
\tilde{\bf Q}(y=+1|I')=&\frac{1}{Z}\exp[\beta S_{top,I'}^{\textrm{new}}|_{\tilde{I}}]\\
S_{top,I'}^{\textrm{new}}|_{\tilde{I}}=&S_{top,I'}+\Delta S_{top,\tilde{I}}e^{-\alpha\cdot dist(I',\tilde{I})}\!\!\!\!\!\!\!\!\!\!
\end{split}
\label{eqn:predict}
\end{equation}
where $S_{top,I'}$ indicates the current AOG's inference score of $S_{top}$ on image $I'$. $S_{top,I'}^{\textrm{new}}|_{\tilde{I}}$ denotes the predicted inference score of $I'$ when people annotate $\tilde{I}$. We assume that if object $I'$ is similar to object $\tilde{I}$, the inference score of $I'$ will have an increase similar to that of $\tilde{I}$. {$\Delta S_{top,\tilde{I}}\!=\!\mathbb{E}_{I\in{\bf I}^{\textrm{ant}}}S_{top,I}-S_{top,\tilde{I}}$} denotes the score increase of $\tilde{I}$. $\alpha$ is a scalar weight. We formulate the appearance distance between $I'$ and $\tilde{I}$ as {$dist(I',\tilde{I})\!=\!1-\frac{\phi(I')^{T}\phi(\tilde{I})}{\vert\phi(I')\vert\cdot\vert\phi(\tilde{I})\vert}$}, where {$\phi(I')\!=\!{\bf M}\,{\bf f}_{I'}$}. ${\bf f}_{I'}$ denotes features of $I'$ at the top conv-layer after ReLU operation, and ${\bf M}$ is a diagonal matrix representing the prior reliability for each feature dimension\footnote{${\bf M}_{ii}\!\propto\!\exp[\mathbb{E}_{I\in{\bf I}}S_{v^{\textrm{unt}}_{i}}]$, where $v^{\textrm{unt}}_{i}$ is the neural unit corresponding to the $i$-th element of ${\bf f}_{I'}$.}. In addition, if $I'$ and $\tilde{I}$ are assigned with different part templates by the current AOG, we set an infinite distance between $I'$ and $\tilde{I}$ to achieve better performance. Based on Equation~(\ref{eqn:predict}), we can predict the changes of the KL divergence after the new annotation on $\tilde{I}$ as
\begin{equation}
\Delta{\bf KL}(\tilde{I})=\lambda{\sum}_{I\in{\bf I}^{\textrm{obj}}}{\sum}_{y}{\bf P}(y|I)\log\frac{\tilde{\bf Q}(y|I)}{{\bf Q}(y|I)}
\label{eqn:delta}
\end{equation}
Thus, in each step, our method selects and asks about the object that decreases the KL divergence the most.
\begin{equation}
\hat{I}={\arg\!\max}_{I\in{\bf I}^{\textrm{unant}}}\Delta{\bf KL}(I)
\label{eqn:select}
\end{equation}

\textbf{QA implementations:}{\verb| |} In the beginning, for each object $I$, we initialize {${\bf P}(y\!=\!+1|I)\!=\!1$} and {${\bf Q}(y\!=\!+1|I)\!=\!0$}. Then, our approach selects and asks about an object $\hat{I}$ based on Equation~(\ref{eqn:select}). We use the answer to update ${\bf P}$. If a new object part is labeled during the QA process, we apply Equation~(\ref{eqn:LossAOG}) to update the AOG. More specifically, if people label a new part template, our method will grow a new AOG branch to encode this template. If people annotate a part for an old part template, our method will update its corresponding AOG branch. Then, we compute the new distribution {${\bf Q}$} based on the new AOG. In this way, the above QA procedure gradually grows the AOG.

\section{Experiments}

\subsection{Implementation details}
\label{sec:implement}

We used a 16-layer VGG network (VGG-16)~\cite{VGG}, which was pre-trained for object classification using 1.3M images in the ImageNet ILSVRC 2012 dataset~\cite{ImageNet}. Then, for each testing category, we further fine-tune the VGG-16 using object images in this category to classify target objects from random images. We selected the last nine conv-layers of VGG-16 as valid conv-layers. We extracted neural units from these conv-layers to build the AOG.

\textbf{Active question-answering:} Three parameters were involved in our active-QA method, \emph{i.e.} $\alpha$, $\beta$, and $Z$. Because most objects of the category contained the target part, we ignored the small probability of ${\bf P}(y=-1|I)$ in Equation~(\ref{eqn:delta}) to simplify the computation. As a result, $Z$ was eliminated in Equation~(\ref{eqn:delta}), and the constant weight $\beta$ did not affect object-selection results in Equation~(\ref{eqn:select}). We set $\alpha=4.0$ in our experiments.

\textbf{Learning AOGs:} Multiple latent patterns corresponding to the same convolutional filter may have similar positions $\overline{\bf p}_{u}$, and their deformation ranges may highly overlap. Thus, we selected the latent pattern with the highest $Score(u)$ within a small range of $\epsilon\times\epsilon$ in the filter's feature map and removed other nearby patterns to obtain a spare AOG. Besides, for each part template $v$, we estimated $n_{k}$ latent patterns in the $k$-th conv-layer. We assumed that scores of all latent patterns in the $k$-th conv-layer follow the distribution of {$Score(u)\sim\alpha\exp[-(\xi\cdot{rank})^{0.5}]+\gamma$}, where $rank$ denotes the score rank of {$u$}. We set {$n_{k}=\lceil0.5/\xi\rceil$}, which learned the best AOG.

\subsection{Datasets}

Because evaluation of part localization requires ground-truth annotations of part positions, we used the following three benchmark datasets to test our method, \emph{i.e.} the PASCAL VOC Part Dataset~\cite{SemanticPart}, the CUB200-2011 dataset~\cite{CUB200}, and the ILSVRC 2013 DET Animal-Part dataset~\cite{CNNAoG}. Just like in \cite{SemanticPart,CNNAoG}, we selected animal categories, which prevalently contain non-rigid shape deformation, for testing. \emph{I.e.} we selected six animal categories---\textit{bird, cat, cow, dog, horse}, and \textit{sheep}---from the PASCAL Part Dataset. The CUB200-2011 dataset contains 11.8K images of 200 bird species. We followed \cite{ActivePart,CNNSemanticPart,CNNAoG} and used all these images as a single bird category for learning. The ILSVRC 2013 DET Animal-Part dataset~\cite{CNNAoG} contains part annotations of 30 animal categories among all the 200 categories in the ILSVRC 2013 DET dataset~\cite{ImageNet}.

\begin{figure*}[t]
\centering
\includegraphics[width=\linewidth]{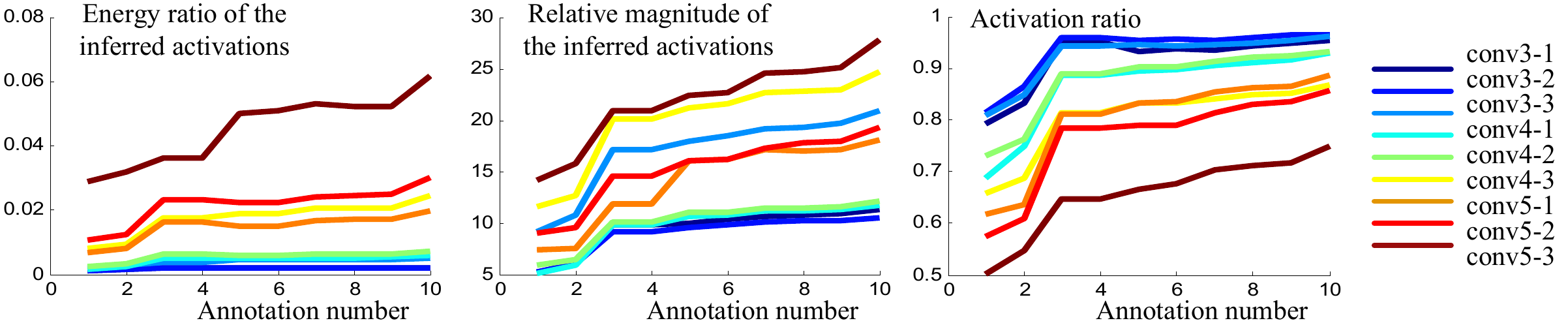}
\caption{Activation states of latent patterns under the selected part template. (left) The ratio of the inferred activation energy to all activation energy in feature maps. (middle) The relative magnitude of the inferred activations, which is normalized by the average activation value of all neural units on the feature map. (right) The ratio of latent patterns that are assigned with an activated neural unit. Different curves shows scores computed based on latent patterns or neural activations in different conv-layers.}
\label{fig:energyCurve}
\end{figure*}

\begin{table}[t]
\begin{center}
\caption{Average number of children of AOG nodes}
\label{tab:stat}
\resizebox{0.75\linewidth}{!}{\begin{tabular}{c|ccc}
\hline
\!\!\!Annotation\!\!\!&\!\!\! \;\;\;Layer 1:\!\!\!&\!\!\! \;\;\;Layer 2:\!\!\!&\!\!\! \;\;\;Layer 3:\!\!\!\\
\!\!\!number\!\!\!&\!\!\!semantic part\!\!\!&\!\!\!part template\!\!\!&\!\!\!latent pattern\!\!\!\\
\hline
\!\!\!{05}
\!\!\!&\!\!\! {3.15} \!\!\!&\!\!\! {3791.5} \!\!\!&\!\!\! {91.6} \!\!\!\\
\hline
\!\!\!{10}
\!\!\!&\!\!\! {5.95} \!\!\!&\!\!\! {3804.8} \!\!\!&\!\!\! {93.9} \!\!\!\\
\hline
\!\!\!{15}
\!\!\!&\!\!\! {8.52} \!\!\!&\!\!\! {3760.4} \!\!\!&\!\!\! {95.5} \!\!\!\\
\hline
\!\!\!{20}
\!\!\!&\!\!\! {11.16} \!\!\!&\!\!\! {3778.3} \!\!\!&\!\!\! {96.3} \!\!\!\\
\hline
\!\!\!{25}
\!\!\!&\!\!\! {13.55} \!\!\!&\!\!\! {3777.5} \!\!\!&\!\!\! {98.3} \!\!\!\\
\hline
\!\!\!{30}
\!\!\!&\!\!\! {15.83} \!\!\!&\!\!\! {3837.3} \!\!\!&\!\!\! {99.2} \!\!\!\\
\hline
\end{tabular}}
\end{center}
\end{table}

\begin{table*}[t]
\begin{center}
\caption{Normalized distance of part localization on the ILSVRC 2013 DET Animal-Part dataset.}
\label{tab:imgnet}
\resizebox{\linewidth}{!}{\begin{tabular}{p{3.3cm}|c|c|cccccccccccccccc}
\hline
\!\!\!&\!\! Part Annot. \!\!\!&\!\! \!{\footnotesize Obj.-box finetune}\! \!\!\!&\!\! gold. \!\!\!&\!\! bird \!\!\!&\!\! frog \!\!\!&\!\! turt. \!\!\!&\!\! liza. \!\!\!&\!\! koala \!\!\!&\!\! lobs. \!\!\!&\!\! dog \!\!\!&\!\! fox \!\!\!&\!\! cat \!\!\!&\!\! lion \!\!\!&\!\! tiger \!\!\!&\!\! bear \!\!\!&\!\! rabb. \!\!\!&\!\! hams. \!\!\!&\!\! squi.\\
\!\!\! SS-DPM-Part~\cite{SSDPM} \!\!\!&\!\! 60 \!\!\!&\!\!  No
\!\!\!&\!\!0.1859
\!\!\!&\!\!0.2747
\!\!\!&\!\!0.2105
\!\!\!&\!\!0.2316
\!\!\!&\!\!0.2901
\!\!\!&\!\!0.1755
\!\!\!&\!\!0.1666
\!\!\!&\!\!0.1948
\!\!\!&\!\!0.1845
\!\!\!&\!\!0.1944
\!\!\!&\!\!0.1334
\!\!\!&\!\!0.0929
\!\!\!&\!\!0.1981
\!\!\!&\!\!0.1355
\!\!\!&\!\!0.1137
\!\!\!&\!\!0.1717
\\
\!\!\! PL-DPM-Part~\cite{PLDPM} \!\!\!&\!\! 60  \!\!\!&\!\! No
\!\!\!&\!\!0.2867
\!\!\!&\!\!0.2337
\!\!\!&\!\!0.2169
\!\!\!&\!\!0.2650
\!\!\!&\!\!0.3079
\!\!\!&\!\!0.1445
\!\!\!&\!\!0.1526
\!\!\!&\!\!0.1904
\!\!\!&\!\!0.2252
\!\!\!&\!\!0.1488
\!\!\!&\!\!0.1450
\!\!\!&\!\!0.1340
\!\!\!&\!\!0.1838
\!\!\!&\!\!0.1968
\!\!\!&\!\!0.1389
\!\!\!&\!\!0.2590
\\
\!\!\! Part-Graph~\cite{SemanticPart} \!\!\!&\!\! 60 \!\!\!&\!\! No
\!\!\!&\!\!0.3385
\!\!\!&\!\!0.3305
\!\!\!&\!\!0.3853
\!\!\!&\!\!0.2873
\!\!\!&\!\!0.3813
\!\!\!&\!\!0.0848
\!\!\!&\!\!0.3467
\!\!\!&\!\!0.1679
\!\!\!&\!\!0.1736
\!\!\!&\!\!0.3499
\!\!\!&\!\!0.1551
\!\!\!&\!\!0.1225
\!\!\!&\!\!0.1906
\!\!\!&\!\!0.2068
\!\!\!&\!\!0.1622
\!\!\!&\!\!0.3038
\\
\!\!\! fc7+linearSVM \!\!\!&\!\! 60 \!\!\!&\!\! Yes
\!\!\!&\!\!0.1359
\!\!\!&\!\!0.2117
\!\!\!&\!\!0.1681
\!\!\!&\!\!0.1890
\!\!\!&\!\!0.2557
\!\!\!&\!\!0.1734
\!\!\!&\!\!0.1845
\!\!\!&\!\!0.1451
\!\!\!&\!\!0.1374
\!\!\!&\!\!0.1581
\!\!\!&\!\!0.1528
\!\!\!&\!\!0.1525
\!\!\!&\!\!0.1354
\!\!\!&\!\!0.1478
\!\!\!&\!\!0.1287
\!\!\!&\!\!0.1291
\\
\!\!\! fc7+RBF-SVM \!\!\!&\!\! 60 \!\!\!&\!\! Yes
\!\!\!&\!\!0.1818
\!\!\!&\!\!0.2637
\!\!\!&\!\!0.2035
\!\!\!&\!\!0.2246
\!\!\!&\!\!0.2538
\!\!\!&\!\!0.1663
\!\!\!&\!\!0.1660
\!\!\!&\!\!0.1512
\!\!\!&\!\!0.1670
\!\!\!&\!\!0.1719
\!\!\!&\!\!0.1176
\!\!\!&\!\!0.1638
\!\!\!&\!\!0.1325
\!\!\!&\!\!0.1312
\!\!\!&\!\!0.1410
\!\!\!&\!\!0.1343
\\
\!\!\! CNN-PDD~\cite{CNNSemanticPart} \!\!\!&\!\! 60 \!\!\!&\!\! No
\!\!\!&\!\!0.1932
\!\!\!&\!\!0.2015
\!\!\!&\!\!0.2734
\!\!\!&\!\!0.2195
\!\!\!&\!\!0.2650
\!\!\!&\!\!0.1432
\!\!\!&\!\!0.1535
\!\!\!&\!\!0.1657
\!\!\!&\!\!0.1510
\!\!\!&\!\!0.1787
\!\!\!&\!\!0.1560
\!\!\!&\!\!0.1756
\!\!\!&\!\!0.1444
\!\!\!&\!\!0.1320
\!\!\!&\!\!0.1251
\!\!\!&\!\!0.1776
\\
\!\!\! CNN-PDD-ft~\cite{CNNSemanticPart} \!\!\!&\!\! 60 \!\!\!&\!\! Yes
\!\!\!&\!\!0.2109
\!\!\!&\!\!0.2531
\!\!\!&\!\!0.1999
\!\!\!&\!\!0.2144
\!\!\!&\!\!0.2494
\!\!\!&\!\!0.1577
\!\!\!&\!\!0.1605
\!\!\!&\!\!0.1847
\!\!\!&\!\!0.1845
\!\!\!&\!\!0.2127
\!\!\!&\!\!0.1521
\!\!\!&\!\!0.2066
\!\!\!&\!\!0.1826
\!\!\!&\!\!0.1595
\!\!\!&\!\!0.1570
\!\!\!&\!\!0.1608
\\
\!\!\! Fast-RCNN (1 ft)~\cite{FastRCNN} \!\!\!&\!\! 30 \!\!\!&\!\! No
\!\!\!&\!\!0.0847
\!\!\!&\!\!0.1520
\!\!\!&\!\!0.1905
\!\!\!&\!\!0.1696
\!\!\!&\!\!0.1412
\!\!\!&\!\!0.0754
\!\!\!&\!\!0.2538
\!\!\!&\!\!0.1471
\!\!\!&\!\!0.0886
\!\!\!&\!\!0.0944
\!\!\!&\!\!0.1004
\!\!\!&\!\!0.0585
\!\!\!&\!\!0.1013
\!\!\!&\!\!0.0821
\!\!\!&\!\!0.0577
\!\!\!&\!\!0.1005
\\
\!\!\! Fast-RCNN (2 fts)~\cite{FastRCNN} \!\!\!&\!\! 30 \!\!\!&\!\! Yes
\!\!\!&\!\!0.0913
\!\!\!&\!\!0.1043
\!\!\!&\!\!0.1294
\!\!\!&\!\!0.1632
\!\!\!&\!\!0.1585
\!\!\!&\!\!0.0730
\!\!\!&\!\!0.2530
\!\!\!&\!\!0.1148
\!\!\!&\!\!0.0736
\!\!\!&\!\!{\bf0.0770}
\!\!\!&\!\!0.0680
\!\!\!&\!\!{\bf0.0441}
\!\!\!&\!\!0.1265
\!\!\!&\!\!0.1017
\!\!\!&\!\!0.0709
\!\!\!&\!\!0.0834
\\
\!\!\! Ours \!\!\!&\!\! \textcolor{red}{\bf 10} \!\!\!&\!\! Yes
\!\!\!&\!\!{\bf 0.0796}
\!\!\!&\!\!{\bf 0.0850}
\!\!\!&\!\!{\bf 0.0906}
\!\!\!&\!\!0.2077
\!\!\!&\!\!{\bf 0.1260}
\!\!\!&\!\!0.0759
\!\!\!&\!\!{\bf 0.1212}
\!\!\!&\!\!0.1476
\!\!\!&\!\!{\bf 0.0584}
\!\!\!&\!\!0.1107
\!\!\!&\!\!0.0716
\!\!\!&\!\!0.0637
\!\!\!&\!\!0.1092
\!\!\!&\!\!{\bf 0.0755}
\!\!\!&\!\!0.0697
\!\!\!&\!\!{\bf 0.0421}
\\
\!\!\! Ours \!\!\!&\!\! \textcolor{red}{\bf 20} \!\!\!&\!\! Yes
\!\!\!&\!\!{\bf 0.0638}
\!\!\!&\!\!{\bf 0.0793}
\!\!\!&\!\!{\bf 0.0765}
\!\!\!&\!\!{\bf 0.1221}
\!\!\!&\!\!{\bf 0.1174}
\!\!\!&\!\!{\bf 0.0720}
\!\!\!&\!\!{\bf 0.1201}
\!\!\!&\!\!{\bf 0.1096}
\!\!\!&\!\!{\bf 0.0517}
\!\!\!&\!\!0.1006
\!\!\!&\!\!0.0752
\!\!\!&\!\!0.0624
\!\!\!&\!\!0.1090
\!\!\!&\!\!{\bf 0.0788}
\!\!\!&\!\!0.0603
\!\!\!&\!\!{\bf 0.0454}
\\
\!\!\! Ours \!\!\!&\!\! \textcolor{red}{\bf 30} \!\!\!&\!\! Yes
\!\!\!&\!\!{\bf 0.0642}
\!\!\!&\!\!{\bf 0.0734}
\!\!\!&\!\!{\bf 0.0971}
\!\!\!&\!\!{\bf 0.0916}
\!\!\!&\!\!{\bf 0.0948}
\!\!\!&\!\!{\bf 0.0658}
\!\!\!&\!\!{\bf 0.1355}
\!\!\!&\!\!{\bf 0.1023}
\!\!\!&\!\!{\bf 0.0474}
\!\!\!&\!\!0.1011
\!\!\!&\!\!{\bf 0.0625}
\!\!\!&\!\!0.0632
\!\!\!&\!\!{\bf 0.0964}
\!\!\!&\!\!{\bf 0.0783}
\!\!\!&\!\!{\bf 0.0540}
\!\!\!&\!\!{\bf 0.0499}
\\
\!\!\!&\!\! \!\!\!&\!\! \!\!\!&\!\! horse \!\!\!&\!\! zebra \!\!\!&\!\! swine \!\!\!&\!\! hippo \!\!\!&\!\! catt. \!\!\!&\!\! sheep \!\!\!&\!\! ante. \!\!\!&\!\! camel \!\!\!&\!\! otter \!\!\!&\!\! arma. \!\!\!&\!\! monk. \!\!\!&\!\! elep. \!\!\!&\!\! red pa. \!\!\!&\!\! gia.pa. \!\!\!&\!\! \!\!\!&\!\! \textcolor{blue}{\bf\large Avg.}\\
\!\!\! SS-DPM-Part~\cite{SSDPM} \!\!\!&\!\! 60 \!\!\!&\!\! No
\!\!\!&\!\!0.2346
\!\!\!&\!\!0.1717
\!\!\!&\!\!0.2262
\!\!\!&\!\!0.2261
\!\!\!&\!\!0.2371
\!\!\!&\!\!0.2364
\!\!\!&\!\!0.2026
\!\!\!&\!\!0.2308
\!\!\!&\!\!0.2088
\!\!\!&\!\!0.2881
\!\!\!&\!\!0.1859
\!\!\!&\!\!0.1740
\!\!\!&\!\!0.1619
\!\!\!&\!\!0.0989
\!\!\!&\!\!
\!\!\!&\!\!\textcolor{blue}{0.1946}
\\
\!\!\! PL-DPM-Part~\cite{PLDPM} \!\!\!&\!\! 60 \!\!\!&\!\! No
\!\!\!&\!\!0.2657
\!\!\!&\!\!0.2937
\!\!\!&\!\!0.2164
\!\!\!&\!\!0.2150
\!\!\!&\!\!0.2320
\!\!\!&\!\!0.2145
\!\!\!&\!\!0.3119
\!\!\!&\!\!0.2949
\!\!\!&\!\!0.2468
\!\!\!&\!\!0.3100
\!\!\!&\!\!0.2113
\!\!\!&\!\!0.1975
\!\!\!&\!\!0.1835
\!\!\!&\!\!0.1396
\!\!\!&\!\!
\!\!\!&\!\!\textcolor{blue}{0.2187}
\\
\!\!\! Part-Graph~\cite{SemanticPart} \!\!\!&\!\! 60 \!\!\!&\!\! No
\!\!\!&\!\!0.2804
\!\!\!&\!\!0.3376
\!\!\!&\!\!0.2979
\!\!\!&\!\!0.2964
\!\!\!&\!\!0.2513
\!\!\!&\!\!0.2321
\!\!\!&\!\!0.3504
\!\!\!&\!\!0.2179
\!\!\!&\!\!0.2535
\!\!\!&\!\!0.2778
\!\!\!&\!\!0.2321
\!\!\!&\!\!0.1961
\!\!\!&\!\!0.1713
\!\!\!&\!\!0.0759
\!\!\!&\!\!
\!\!\!&\!\!\textcolor{blue}{0.2486}
\\
\!\!\! fc7+linearSVM \!\!\!&\!\! 60 \!\!\!&\!\! Yes
\!\!\!&\!\!0.2003
\!\!\!&\!\!0.2409
\!\!\!&\!\!0.1632
\!\!\!&\!\!0.1400
\!\!\!&\!\!0.2043
\!\!\!&\!\!0.2274
\!\!\!&\!\!0.1479
\!\!\!&\!\!0.2204
\!\!\!&\!\!0.2498
\!\!\!&\!\!0.2875
\!\!\!&\!\!0.2261
\!\!\!&\!\!0.1520
\!\!\!&\!\!0.1557
\!\!\!&\!\!0.1071
\!\!\!&\!\!
\!\!\!&\!\!\textcolor{blue}{0.1776}
\\
\!\!\! fc7+RBF-SVM \!\!\!&\!\! 60 \!\!\!&\!\! Yes
\!\!\!&\!\!0.2207
\!\!\!&\!\!0.1550
\!\!\!&\!\!0.1963
\!\!\!&\!\!0.1536
\!\!\!&\!\!0.2609
\!\!\!&\!\!0.2295
\!\!\!&\!\!0.1748
\!\!\!&\!\!0.2080
\!\!\!&\!\!0.2263
\!\!\!&\!\!0.2613
\!\!\!&\!\!0.2244
\!\!\!&\!\!0.1806
\!\!\!&\!\!0.1417
\!\!\!&\!\!0.1095
\!\!\!&\!\!
\!\!\!&\!\!\textcolor{blue}{0.1838}
\\
\!\!\! CNN-PDD~\cite{CNNSemanticPart} \!\!\!&\!\! 60 \!\!\!&\!\! No
\!\!\!&\!\!0.2610
\!\!\!&\!\!0.2363
\!\!\!&\!\!0.1623
\!\!\!&\!\!0.2018
\!\!\!&\!\!0.1955
\!\!\!&\!\!0.1350
\!\!\!&\!\!0.1857
\!\!\!&\!\!0.2499
\!\!\!&\!\!0.2486
\!\!\!&\!\!0.2656
\!\!\!&\!\!0.1704
\!\!\!&\!\!0.1765
\!\!\!&\!\!0.1713
\!\!\!&\!\!0.1638
\!\!\!&\!\!
\!\!\!&\!\!\textcolor{blue}{0.1893}
\\
\!\!\! CNN-PDD-ft~\cite{CNNSemanticPart} \!\!\!&\!\! 60 \!\!\!&\!\! Yes
\!\!\!&\!\!0.2417
\!\!\!&\!\!0.2725
\!\!\!&\!\!0.1943
\!\!\!&\!\!0.2299
\!\!\!&\!\!0.2104
\!\!\!&\!\!0.1936
\!\!\!&\!\!0.1712
\!\!\!&\!\!0.2552
\!\!\!&\!\!0.2110
\!\!\!&\!\!0.2726
\!\!\!&\!\!0.1463
\!\!\!&\!\!0.1602
\!\!\!&\!\!0.1868
\!\!\!&\!\!0.1475
\!\!\!&\!\!
\!\!\!&\!\!\textcolor{blue}{0.1980}
\\
\!\!\! Fast-RCNN (1 ft)~\cite{FastRCNN} \!\!\!&\!\! 30 \!\!\!&\!\! No
\!\!\!&\!\!0.2694
\!\!\!&\!\!{\bf0.0823}
\!\!\!&\!\!0.1319
\!\!\!&\!\!0.0976
\!\!\!&\!\!0.1309
\!\!\!&\!\!0.1276
\!\!\!&\!\!0.1348
\!\!\!&\!\!0.1609
\!\!\!&\!\!0.1627
\!\!\!&\!\!0.1889
\!\!\!&\!\!0.1367
\!\!\!&\!\!{\bf0.1081}
\!\!\!&\!\!0.0791
\!\!\!&\!\!{\bf0.0474}
\!\!\!&\!\!
\!\!\!&\!\!\textcolor{blue}{0.1252}
\\
\!\!\! Fast-RCNN (2 fts)~\cite{FastRCNN} \!\!\!&\!\! 30 \!\!\!&\!\! Yes
\!\!\!&\!\!0.1629
\!\!\!&\!\!0.0881
\!\!\!&\!\!{\bf0.1228}
\!\!\!&\!\!{\bf0.0889}
\!\!\!&\!\!{\bf0.0922}
\!\!\!&\!\!{\bf0.0622}
\!\!\!&\!\!0.1000
\!\!\!&\!\!0.1519
\!\!\!&\!\!0.0969
\!\!\!&\!\!{\bf0.1485}
\!\!\!&\!\!0.0855
\!\!\!&\!\!0.1085
\!\!\!&\!\!{\bf0.0407}
\!\!\!&\!\!0.0542
\!\!\!&\!\!
\!\!\!&\!\!\textcolor{blue}{0.1045}
\\
\!\!\! Ours \!\!\!&\!\! \textcolor{red}{\bf 10} \!\!\!&\!\! Yes
\!\!\!&\!\!{\bf 0.1297}
\!\!\!&\!\!0.1413
\!\!\!&\!\!0.2145
\!\!\!&\!\!0.1377
\!\!\!&\!\!0.1493
\!\!\!&\!\!0.1415
\!\!\!&\!\!0.1046
\!\!\!&\!\!{\bf 0.1239}
\!\!\!&\!\!0.1288
\!\!\!&\!\!0.1964
\!\!\!&\!\!{\bf 0.0524}
\!\!\!&\!\!0.1507
\!\!\!&\!\!0.1081
\!\!\!&\!\!0.0640
\!\!\!&\!\!
\!\!\!&\!\!\textcolor{blue}{0.1126}
\\
\!\!\! Ours \!\!\!&\!\! \textcolor{red}{\bf 20} \!\!\!&\!\! Yes
\!\!\!&\!\!{\bf 0.1083}
\!\!\!&\!\!0.1389
\!\!\!&\!\!0.1475
\!\!\!&\!\!0.1280
\!\!\!&\!\!0.1490
\!\!\!&\!\!0.1300
\!\!\!&\!\!{\bf 0.0667}
\!\!\!&\!\!{\bf 0.1033}
\!\!\!&\!\!0.1103
\!\!\!&\!\!0.1526
\!\!\!&\!\!{\bf 0.0497}
\!\!\!&\!\!0.1301
\!\!\!&\!\!0.0802
\!\!\!&\!\!0.0574
\!\!\!&\!\!
\!\!\!&\!\!\textcolor{blue}{\bf 0.0965}
\\
\!\!\! Ours \!\!\!&\!\! \textcolor{red}{\bf 30} \!\!\!&\!\! Yes
\!\!\!&\!\!{\bf 0.1129}
\!\!\!&\!\!0.1066
\!\!\!&\!\!0.1408
\!\!\!&\!\!0.1204
\!\!\!&\!\!0.1118
\!\!\!&\!\!0.1260
\!\!\!&\!\!{\bf 0.0825}
\!\!\!&\!\!{\bf 0.0836}
\!\!\!&\!\!{\bf 0.0901}
\!\!\!&\!\!0.1685
\!\!\!&\!\!{\bf 0.0490}
\!\!\!&\!\!0.1224
\!\!\!&\!\!0.0779
\!\!\!&\!\!0.0577
\!\!\!&\!\!
\!\!\!&\!\!\textcolor{blue}{\bf 0.0909}
\\
\hline
\end{tabular}}
\end{center}
\vspace{1pt}
\noindent The 2nd column shows the number of part annotations for training. The 3rd column indicates whether the baseline used all object-box annotations in the category to pre-fine-tune a CNN before learning the part (\textit{object-box annotations are more than part annotations}).
\end{table*}

\subsection{Baselines}

We used the following thirteen baselines for comparison. The first two baselines were based on the Fast-RCNN~\cite{FastRCNN}. We fine-tuned the fast-RCNN with a loss of detecting a single class/part for a fair comparison. The first baseline, namely \textit{Fast-RCNN (1 ft)}, fine-tuned the VGG-16 using part annotations to detect parts on well-cropped objects. To enable a more fair comparison, we conducted the second baseline based on two-stage fine-tuning, namely \textit{Fast-RCNN (2 fts)}. This baseline first fine-tuned the VGG-16 using numerous object-box annotations in the target category, and then fine-tuned the VGG-16 using a few part annotations.

The third baseline was proposed in \cite{CNNSemanticPart}, namely \textit{CNN-PDD}. \textit{CNN-PDD} selected a filter in a CNN (pre-trained using ImageNet ILSVRC 2012 dataset) to represent the part on well-cropped objects. Then, we slightly extended \cite{CNNSemanticPart} as the fourth baseline \textit{CNN-PDD-ft}. \textit{CNN-PDD-ft} first fine-tuned the VGG-16 using object bounding boxes, and then applied \cite{CNNSemanticPart} to learn object parts.

The strongly supervised DPM (\textit{SS-DPM-Part})~\cite{SSDPM} and the approach of \cite{PLDPM} (\textit{PL-DPM-Part}) were the fifth and sixth baselines. These methods learned DPMs for part localization. The graphical model proposed in \cite{SemanticPart} was selected as the seventh baseline, namely \textit{Part-Graph}. The eighth baseline was the interactive learning for part localization~\cite{ActivePart} (\textit{Interactive-DPM}).

Without lots of training samples, ``simple'' methods are usually insensitive to the over-fitting problem. Thus, we designed the last four baselines as follows. We first fine-tuned the VGG-16 using object bounding boxes, and collected image patches from cropped objects based on the selective search~\cite{SelectiveSearch}. We used the VGG-16 to extract \textit{fc7} features from image patches. The two baselines (\emph{i.e.} \textit{fc7+linearSVM} and \textit{fc7+RBF-SVM}) used a linear SVM and an RBF-SVM, respectively, to detect object parts. The other baselines \textit{VAE+linearSVM} and \textit{CoopNet+linearSVM} used features of the VAE network~\cite{VAE} and the CoopNet~\cite{CoopNet}, respectively, instead of \textit{fc7} features, for part detection.

The last baseline~\cite{CNNAoG} learned AOGs without QA (\textit{AOG w/o QA}). We randomly selected objects and annotated their parts for training.

Both object annotations and part annotations are used to learn models in all the thirteen baselines (including those without fine-tuning). \textit{Fast-RCNN (1 ft)} and \textit{CNN-PDD} used the cropped objects as the input of the CNN; \textit{SS-DPM-Part}, \textit{PL-DPM-Part}, \textit{Part-Graph}, and \textit{Interactive-DPM} used object boxes and part boxes to learn models. \textit{CNN-PDD-ft}, \textit{Fast-RCNN (2 fts)}, and methods based on \textit{fc7} features used object bounding boxes for fine-tuning.

\subsection{Evaluation metric}

As discussed in \cite{SemanticPart,CNNAoG}, a fair evaluation of part localization requires removing factors of object detection. Thus, we used ground-truth object bounding boxes to crop objects as testing images. Given an object image, some competing methods (\emph{e.g.} \textit{Fast-RCNN (1 ft)}, \textit{Part-Graph}, and \textit{SS-DPM-Part}) estimate several bounding boxes for the part with different confidences. We followed \cite{CNNSemanticPart,SemanticPart,ObjectDiscoveryCNN_1,CNNAoG} to take the most confident bounding box per image as the part-localization result. Given part-localization results of a category, we applied the \textit{normalized distance}~\cite{CNNSemanticPart} and the \textit{percentage of correctly localized parts} (PCP)~\cite{fineGrained1,fineGrained2,fineGrained3} to evaluate the localization accuracy. We measured the distance between the predicted part center and the ground-truth part center, and then normalized the distance using the diagonal length of the object as the normalized distance. For the PCP, we used the typical metric of ``$IoU\geq0.5$''~\cite{FastRCNN} to identify correct part localizations.

\begin{table}[t]
\begin{center}
\caption{Part localization performance on the CUB200 dataset.}
\label{tab:cub200}
\resizebox{\linewidth}{!}{\begin{tabular}{lcccc}
\hline
\!\!\!&\!\! \!\!\!\!\!\!\!\!\!\!\!\!{\small Obj.-box finetune} \!\!\!&\!\! {\small Part Annot.} \!\!\!&\!\! \!\#Q \!\!\!&\!\!{\small Normalizaed distance}\!\!\!\\
\!\!\! SS-DPM-Part~\cite{SSDPM} \!\!\!&\!\! No \!\!\!&\!\! 60 \!\!\!&\!\! -- \!\!\!& 0.2504\\
\!\!\! PL-DPM-Part~\cite{PLDPM} \!\!\!&\!\! No \!\!\!&\!\! 60 \!\!\!&\!\! -- \!\!\!& 0.3215\\
\!\!\! Part-Graph~\cite{SemanticPart} \!\!\!&\!\! No \!\!\!&\!\! 60 \!\!\!&\!\! -- \!\!\!& 0.3697\\
\!\!\! fc7+linearSVM \!\!\!&\!\! Yes \!\!\!&\!\! 60 \!\!\!&\!\! -- \!\!\!& 0.2786\\
\!\!\! fc7+RBF-SVM \!\!\!&\!\! Yes \!\!\!&\!\! 60 \!\!\!&\!\! -- \!\!\!& 0.3360\\
\!\!\! Interactive-DPM~\cite{ActivePart} \!\!\!&\!\! No \!\!\!&\!\! 60 \!\!\!&\!\! -- \!\!\! & 0.2011\\
\!\!\! CNN-PDD~\cite{CNNSemanticPart} \!\!\!&\!\! No \!\!\!&\!\! 60 \!\!\!&\!\! -- \!\!\! & 0.2446\\
\!\!\! CNN-PDD-ft~\cite{CNNSemanticPart} \!\!\!&\!\! Yes \!\!\!&\!\! 60 \!\!\!&\!\! -- \!\!\! & 0.2694\\
\!\!\! Fast-RCNN (1 ft)~\cite{FastRCNN} \!\!\!&\!\! No \!\!\!&\!\! 60 \!\!\!&\!\! -- \!\!\! & 0.3105\\
\!\!\! Fast-RCNN (2 fts)~\cite{FastRCNN} \!\!\!&\!\! Yes \!\!\!&\!\! 60 \!\!\!&\!\! -- \!\!\! & 0.1989\\
\!\!\! AOG w/o QA~\cite{CNNAoG} \!\!\!&\!\! Yes \!\!\!&\!\! \textcolor{red}{\bf 20} \!\!\!&\!\! -- \!\!\! & 0.1084\\
\!\!\! Ours \!\!\!&\!\! Yes \!\!\!&\!\! \textcolor{red}{\bf 10} \!\!\!&\!\! 28 \!\!\! & {\bf 0.0626}\\
\!\!\! Ours \!\!\!&\!\! Yes \!\!\!&\!\! \textcolor{red}{\bf 20} \!\!\!&\!\! 112 \!\!\! & {\bf 0.0434}\\
\hline
\end{tabular}}
\end{center}
\vspace{1pt}
\noindent See Table~\ref{tab:imgnet} for the introduction of the 2nd and 3rd columns. The 4th column shows the number of questions for training. The 4th column indicates whether the baseline used all object annotations (\textit{more than part annotations}) in the category to pre-fine-tune a CNN before learning the part.
\end{table}

\begin{table}[t]
\begin{center}
\caption{Part localization on the Pascal VOC Part dataset.}
\label{tab:VOC}
\resizebox{\linewidth}{!}{\begin{tabular}{c|lccccccccc}
\hline
\!\!\! & \!\!\! Method \!\!\!&\!\! \!\!\!\!\!\!Annot.\!\! \!\!\!&\!\! \#Q \!\!\!&\!\! bird \!\!\!&\!\! cat \!\!\!&\!\! cow \!\!\!&\!\! dog \!\!\!&\!\! {\small horse} \!\!\!&\!\! {\small sheep} \!\!\!&\!\! \textcolor{blue}{\bf Avg.}\!\!\!\\
\hline
\multirow{7}{*}{\rotatebox[origin=c]{90}{Head}} &\!\!\! {\small Fast-RCNN (1 ft)~\cite{FastRCNN}} \!\!\!&\!\! \textcolor{red}{\bf 10} \!\!\!&\!\! --
\!\!\!&\!\!0.326
\!\!\!&\!\!0.238
\!\!\!&\!\!0.283
\!\!\!&\!\!0.286
\!\!\!&\!\!0.319
\!\!\!&\!\!0.354
\!\!\!&\!\!\textcolor{blue}{0.301}\!\!\!
\\
& \!\!\! {\small Fast-RCNN (2 fts)~\cite{FastRCNN}} \!\!\!&\!\! \textcolor{red}{\bf 10} \!\!\!&\!\! --
\!\!\!&\!\!0.233
\!\!\!&\!\!0.196
\!\!\!&\!\!0.216
\!\!\!&\!\!0.206
\!\!\!&\!\!0.253
\!\!\!&\!\!0.286
\!\!\!&\!\!\textcolor{blue}{0.232}\!\!\!
\\
& \!\!\! {\small Fast-RCNN (1 ft)~\cite{FastRCNN}} \!\!\!&\!\! 20 \!\!\!&\!\! --
\!\!\!&\!\!0.352
\!\!\!&\!\!{\bf0.131}
\!\!\!&\!\!0.275
\!\!\!&\!\!0.189
\!\!\!&\!\!0.293
\!\!\!&\!\!0.252
\!\!\!&\!\!\textcolor{blue}{0.249}\!\!\!
\\
& \!\!\! {\small Fast-RCNN (2 fts)~\cite{FastRCNN}} \!\!\!&\!\! 20 \!\!\!&\!\! --
\!\!\!&\!\!0.176
\!\!\!&\!\!0.132
\!\!\!&\!\!0.191
\!\!\!&\!\!0.171
\!\!\!&\!\!0.231
\!\!\!&\!\!{\bf0.189}
\!\!\!&\!\!\textcolor{blue}{0.182}\!\!\!
\\
& \!\!\! {\small Fast-RCNN (1 ft)~\cite{FastRCNN}} \!\!\!&\!\! 30 \!\!\!&\!\! --
\!\!\!&\!\!0.285
\!\!\!&\!\!0.146
\!\!\!&\!\!0.228
\!\!\!&\!\!0.141
\!\!\!&\!\!0.250
\!\!\!&\!\!0.220
\!\!\!&\!\!\textcolor{blue}{0.212}\!\!\!
\\
& \!\!\! {\small Fast-RCNN (2 fts)~\cite{FastRCNN}} \!\!\!&\!\! 30 \!\!\!&\!\! --
\!\!\!&\!\!0.173
\!\!\!&\!\!0.156
\!\!\!&\!\!0.150
\!\!\!&\!\!{\bf0.137}
\!\!\!&\!\!0.132
\!\!\!&\!\!0.221
\!\!\!&\!\!\textcolor{blue}{0.161}\!\!\!
\\
& \!\!\! {\small Ours} \!\!\!&\!\! \textcolor{red}{\bf 10} \!\!\!&\!\! 14.7
\!\!\!&\!\!{\bf0.144}
\!\!\!&\!\!0.146
\!\!\!&\!\!{\bf0.137}
\!\!\!&\!\!0.145
\!\!\!&\!\!{\bf0.122}
\!\!\!&\!\!0.193
\!\!\!&\!\!\textcolor{blue}{\bf0.148}\!\!\!
\\
\hline
\multirow{7}{*}{\rotatebox[origin=c]{90}{Neck}} &\!\!\! {\small Fast-RCNN (1 ft)~\cite{FastRCNN}} \!\!\!&\!\! \textcolor{red}{\bf 10} \!\!\!&\!\! --
\!\!\!&\!\!0.251
\!\!\!&\!\!0.333
\!\!\!&\!\!0.310
\!\!\!&\!\!0.248
\!\!\!&\!\!0.267
\!\!\!&\!\!0.242
\!\!\!&\!\!\textcolor{blue}{0.275}\!\!\!
\\
& \!\!\! {\small Fast-RCNN (2 fts)~\cite{FastRCNN}} \!\!\!&\!\! \textcolor{red}{\bf 10} \!\!\!&\!\! --
\!\!\!&\!\!0.317
\!\!\!&\!\!0.335
\!\!\!&\!\!0.307
\!\!\!&\!\!0.362
\!\!\!&\!\!0.271
\!\!\!&\!\!0.259
\!\!\!&\!\!\textcolor{blue}{0.309}\!\!\!
\\
& \!\!\! {\small Fast-RCNN (1 ft)~\cite{FastRCNN}} \!\!\!&\!\! 20 \!\!\!&\!\! --
\!\!\!&\!\!0.255
\!\!\!&\!\!0.359
\!\!\!&\!\!0.241
\!\!\!&\!\!0.281
\!\!\!&\!\!0.268
\!\!\!&\!\!0.235
\!\!\!&\!\!\textcolor{blue}{0.273}\!\!\!
\\
& \!\!\! {\small Fast-RCNN (2 fts)~\cite{FastRCNN}} \!\!\!&\!\! 20 \!\!\!&\!\! --
\!\!\!&\!\!0.260
\!\!\!&\!\!0.289
\!\!\!&\!\!0.304
\!\!\!&\!\!0.297
\!\!\!&\!\!0.255
\!\!\!&\!\!0.237
\!\!\!&\!\!\textcolor{blue}{0.274}\!\!\!
\\
& \!\!\! {\small Fast-RCNN (1 ft)~\cite{FastRCNN}} \!\!\!&\!\! 30 \!\!\!&\!\! --
\!\!\!&\!\!0.288
\!\!\!&\!\!0.324
\!\!\!&\!\!0.247
\!\!\!&\!\!0.262
\!\!\!&\!\!0.210
\!\!\!&\!\!0.220
\!\!\!&\!\!\textcolor{blue}{0.258}\!\!\!
\\
& \!\!\! {\small Fast-RCNN (2 fts)~\cite{FastRCNN}} \!\!\!&\!\! 30 \!\!\!&\!\! --
\!\!\!&\!\!0.201
\!\!\!&\!\!0.276
\!\!\!&\!\!0.281
\!\!\!&\!\!0.254
\!\!\!&\!\!0.220
\!\!\!&\!\!0.229
\!\!\!&\!\!\textcolor{blue}{0.244}\!\!\!
\\
& \!\!\! {\small Ours} \!\!\!&\!\! \textcolor{red}{\bf 10} \!\!\!&\!\! 24.5
\!\!\!&\!\!{\bf0.120}
\!\!\!&\!\!{\bf0.144}
\!\!\!&\!\!{\bf0.178}
\!\!\!&\!\!{\bf0.152}
\!\!\!&\!\!{\bf0.161}
\!\!\!&\!\!{\bf0.161}
\!\!\!&\!\!\textcolor{blue}{\bf0.152}\!\!\!
\\
\hline
\multirow{7}{*}{\rotatebox[origin=c]{90}{\scriptsize Nose/Muzzle/Beek}} &\!\!\! {\small Fast-RCNN (1 ft)~\cite{FastRCNN}} \!\!\!&\!\! \textcolor{red}{\bf 10} \!\!\!&\!\! --
\!\!\!&\!\!0.446
\!\!\!&\!\!0.389
\!\!\!&\!\!0.301
\!\!\!&\!\!0.326
\!\!\!&\!\!0.385
\!\!\!&\!\!0.328
\!\!\!&\!\!\textcolor{blue}{0.363}\!\!\!
\\
& \!\!\! {\small Fast-RCNN (2 fts)~\cite{FastRCNN}} \!\!\!&\!\! \textcolor{red}{\bf 10} \!\!\!&\!\! --
\!\!\!&\!\!0.447
\!\!\!&\!\!0.433
\!\!\!&\!\!0.313
\!\!\!&\!\!0.391
\!\!\!&\!\!0.338
\!\!\!&\!\!0.350
\!\!\!&\!\!\textcolor{blue}{0.379}\!\!\!
\\
& \!\!\! {\small Fast-RCNN (1 ft)~\cite{FastRCNN}} \!\!\!&\!\! 20 \!\!\!&\!\! --
\!\!\!&\!\!0.425
\!\!\!&\!\!0.372
\!\!\!&\!\!0.260
\!\!\!&\!\!0.303
\!\!\!&\!\!0.334
\!\!\!&\!\!0.279
\!\!\!&\!\!\textcolor{blue}{0.329}\!\!\!
\\
& \!\!\! {\small Fast-RCNN (2 fts)~\cite{FastRCNN}} \!\!\!&\!\! 20 \!\!\!&\!\! --
\!\!\!&\!\!0.419
\!\!\!&\!\!0.351
\!\!\!&\!\!0.289
\!\!\!&\!\!0.249
\!\!\!&\!\!0.296
\!\!\!&\!\!0.293
\!\!\!&\!\!\textcolor{blue}{0.316}\!\!\!
\\
& \!\!\! {\small Fast-RCNN (1 ft)~\cite{FastRCNN}} \!\!\!&\!\! 30 \!\!\!&\!\! --
\!\!\!&\!\!0.462
\!\!\!&\!\!0.336
\!\!\!&\!\!0.242
\!\!\!&\!\!0.260
\!\!\!&\!\!0.247
\!\!\!&\!\!0.257
\!\!\!&\!\!\textcolor{blue}{0.301}\!\!\!
\\
& \!\!\! {\small Fast-RCNN (2 fts)~\cite{FastRCNN}} \!\!\!&\!\! 30 \!\!\!&\!\! --
\!\!\!&\!\!0.430
\!\!\!&\!\!0.338
\!\!\!&\!\!0.239
\!\!\!&\!\!0.219
\!\!\!&\!\!0.271
\!\!\!&\!\!0.285
\!\!\!&\!\!\textcolor{blue}{0.297}\!\!\!
\\
& \!\!\! {\small Ours} \!\!\!&\!\! \textcolor{red}{\bf 10} \!\!\!&\!\! 23.8
\!\!\!&\!\!{\bf0.134}
\!\!\!&\!\!{\bf0.112}
\!\!\!&\!\!{\bf0.182}
\!\!\!&\!\!{\bf0.156}
\!\!\!&\!\!{\bf0.217}
\!\!\!&\!\!{\bf0.181}
\!\!\!&\!\!\textcolor{blue}{\bf0.164}\!\!\!
\\
\hline
\end{tabular}}
\end{center}
\vspace{1pt}
\noindent The 3rd and 4th columns show the number of part annotations and the average number of questions for training.
\end{table}

\subsection{Experimental results}

We learned AOGs for the head, the neck, and the nose/muzzle/beak parts of the six animal categories in the Pascal VOC Part dataset. For the ILSVRC 2013 DET Animal-Part dataset and the CUB200-2011 dataset, we learned an AOG for the head part\footnote{It is the ``forehead'' part for birds in the CUB200-2011 dataset.} of each category. It is because all categories in the two datasets contain the head part. We did not train human annotators. Shape differences between two part templates were often very vague, so that an annotator could assign a part to either part template.

\begin{figure*}[t]
\centering
\includegraphics[width=\linewidth]{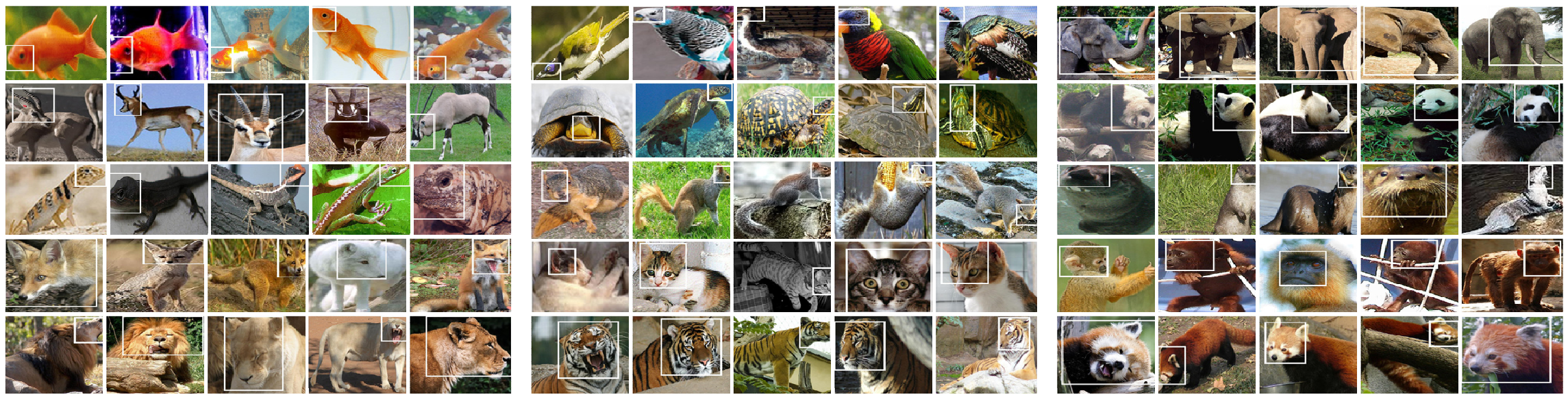}
\caption{Part localization results based on AOGs.}
\label{fig:results}
\includegraphics[width=\linewidth]{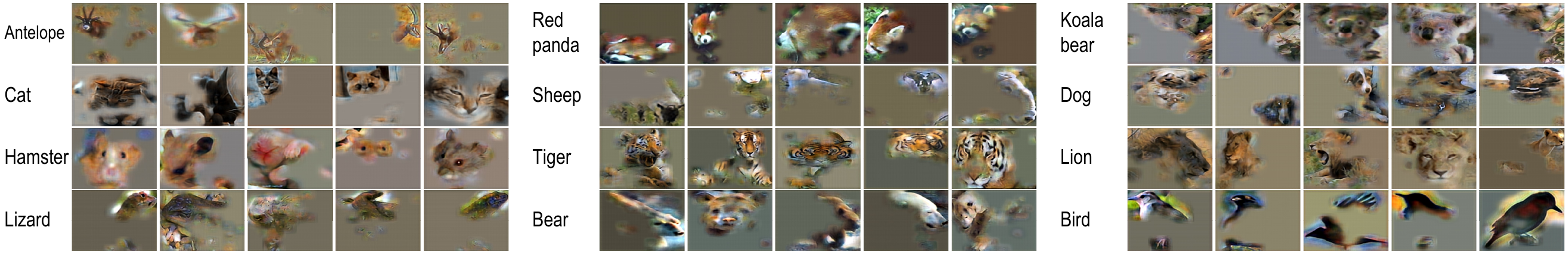}
\caption{Visualization of latent patterns in AOGs for the head part. The up-convolutional net~\cite{FeaVisual} synthesizes images corresponding neural activations, which are selected by the AOG during part parsing. We only visualize neural activations selected from conv-layers 5--7. Some latent patterns select neural units corresponding to constituent regions \emph{w.r.t.} the target part, while other latent patterns describe contexts.}
\label{fig:visualization}
\end{figure*}

\begin{figure}[t]
\centering
\includegraphics[width=0.8\linewidth]{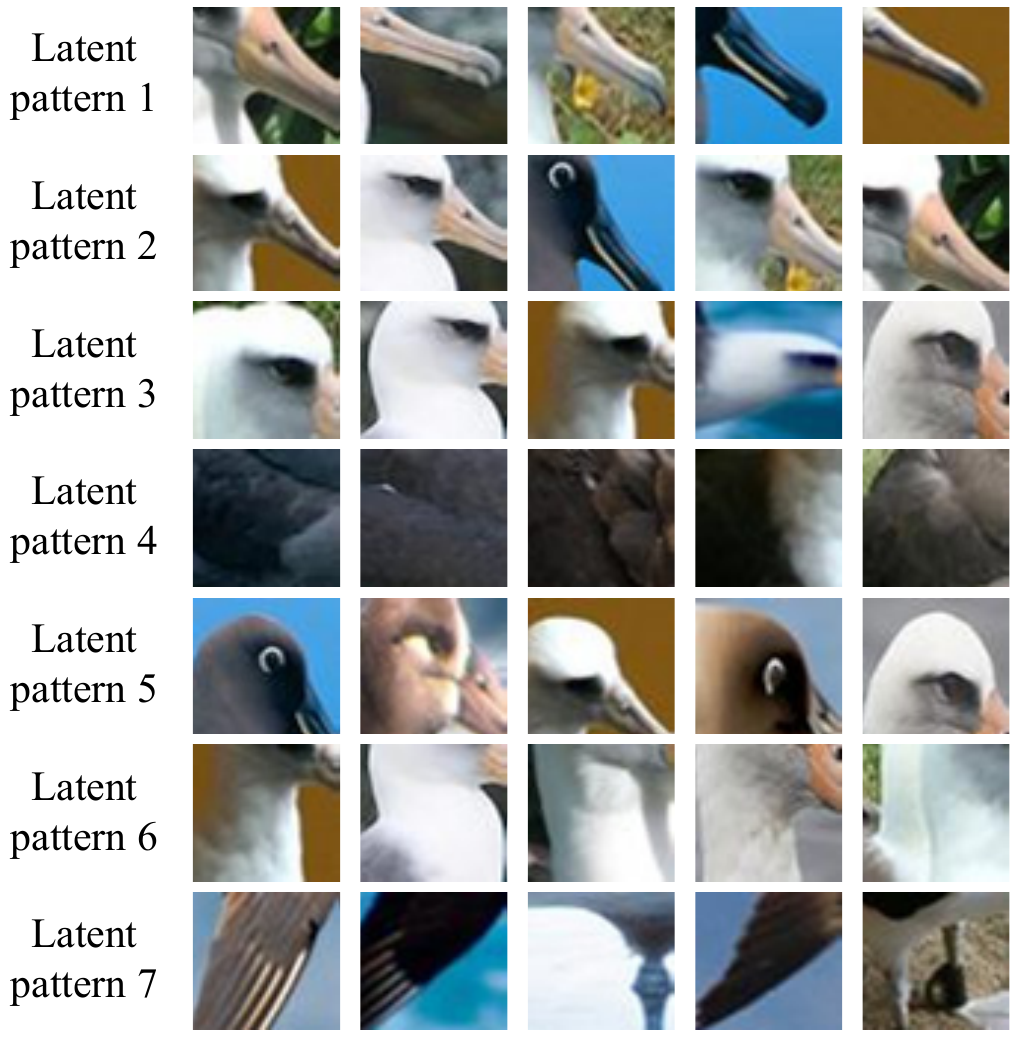}
\caption{Image patches corresponding to different latent patterns.}
\label{fig:patches}
\end{figure}

Table~\ref{tab:stat} shows how the AOG grew when people annotated more parts during the QA process. Given AOGs learned for the PASCAL VOC Part dataset, we computed the average number of children for each node in different AOG layers. The AOG mainly grew by adding new branches to represent new part templates. The refinement of an existing AOG branch did not significantly change the node number of the AOG.

Fig.~\ref{fig:energyCurve} analyzes activation states of latent patterns in AOGs that were learned with different numbers of part annotations. Given a testing image $I$ for part parsing, we only focused on the inferred latent patterns and neural units, \emph{i.e.} latent patterns and their inferred neural units under the selected part template. Let ${\bf V}$ and ${\bf V'}\subset{\bf V}$ denote all units in a specific conv-layer and the inferred units, respectively. $a_{v}$ denotes the activation score of $v\in{\bf V}$ after the ReLU operation. $a_{v}$ is also normalized by the average activation level of $v$'s corresponding feature maps \emph{w.r.t.} different images. Thus, in Fig.~\ref{fig:energyCurve}(left), we computed the ratio of the inferred activation energy as $\frac{\sum_{v\in{\bf V'}}a_{v}}{\sum_{v\in{\bf V}}a_{v}}$. For each inferred latent pattern $u$, $a_{u}$ denotes the activation score of its selected neural unit\footnote{Two latent patterns may select the same neural unit}. Fig.~\ref{fig:energyCurve}(middle) measures the relative magnitude of the inferred activations, which was measured as $\frac{\mathbb{E}_{u\in{\bf U}}[a_{u}]}{\mathbb{E}_{v\in{\bf V}}[a_{v}]}$. Fig.~\ref{fig:energyCurve}(right) shows the ratio of the latent patterns being strongly activated. We used a threshold $\tau=\mathbb{E}_{v\in{\bf V}}[a_{v}]$ to identify strong activations, \emph{i.e.} computing the activation ratio as $\mathbb{E}_{u\in{\bf U}}[{\bf 1}(a_{u}>\tau)]$. Curves in Fig.~\ref{fig:energyCurve} were reported as the average performance using images in the CUB200-2011 dataset.

Fig.~\ref{fig:visualization} visualizes latent patterns in the AOG based on the technique of \cite{FeaVisual}. More specifically, Fig.~\ref{fig:patches} lists images patches inferred by different latent patterns in the AOG with high inference scores. It shows that each latent pattern corresponds to a specific part shape through different images.

\begin{figure}[t]
\centering
\includegraphics[width=\linewidth]{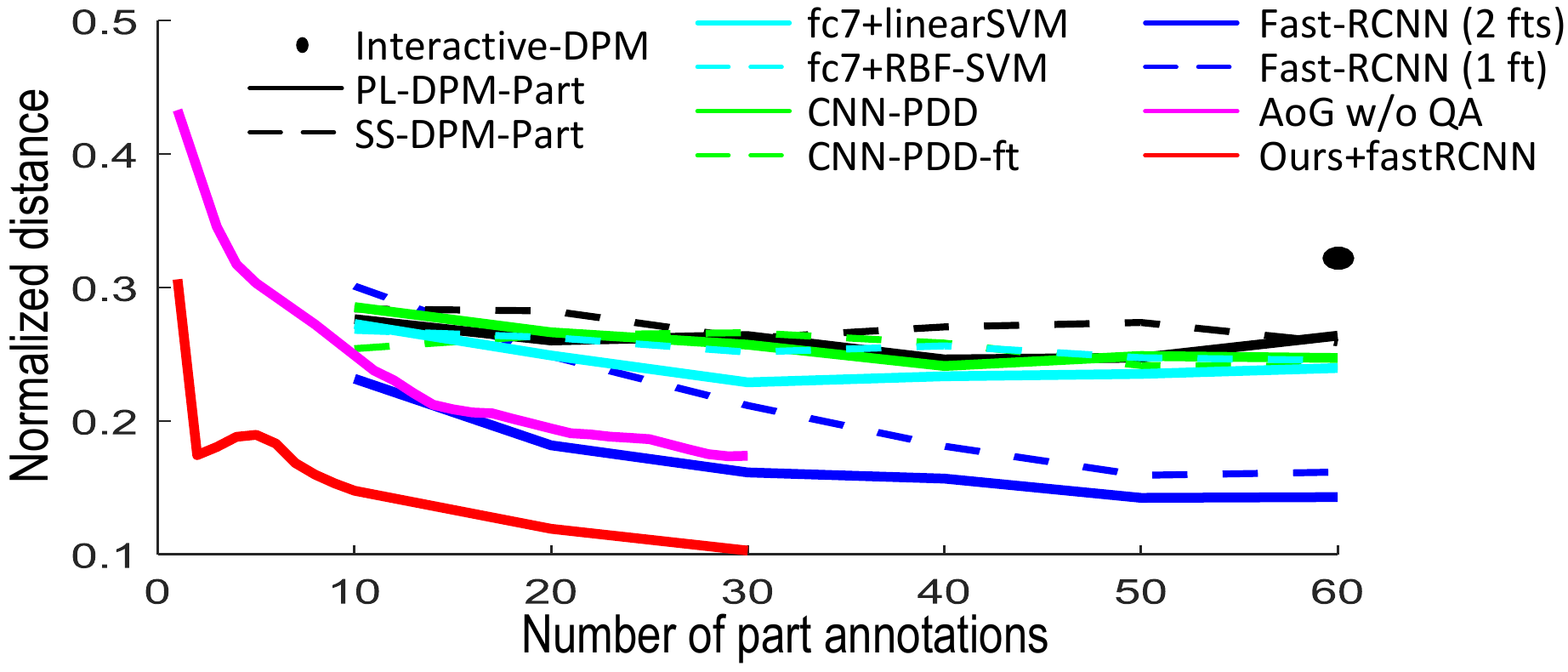}
\caption{Part localization performance on the Pascal VOC Part dataset.}
\label{fig:curve}
\end{figure}

Fig.~\ref{fig:results} shows part localization results based on AOGs. Tables~\ref{tab:imgnet}, \ref{tab:VOC}, and \ref{tab:cub200} compare the part-localization performance of different baselines on different benchmark datasets using the evaluation metric of the normalized distance. Tables~\ref{tab:VOC}, and \ref{tab:cub200} show both the number of part annotations and the number of questions. Fig.~\ref{fig:curve} shows the performance of localizing the head part on objects in the PASCAL VOC Part Dataset, when people annotated different numbers of parts for training. Table~\ref{tab:pcp} lists part-localization performance, which was evaluated by the PCP metric. In particular, the method of \textit{Ours+fastRCNN} combined our method and the fast-RCNN to refine part-localization results\footnote{We used part boxes annotated during the QA process to learn a fast-RCNN for part detection. Given the inference result $\Lambda_{v}$ of part template $v$ on image $I$, we define a new inference score for localization refinement $S_{v}^{\textrm{new}}(\Lambda_{v}^{\textrm{new}})=S_{v}+\lambda_1\Phi(\Lambda^{\textrm{new}}_{v})+\lambda_2\frac{\Vert{\bf p}_{v}-{\bf p}_{v}^{\textrm{new}}\Vert}{2\sigma^2}$, where $\sigma=70$ pixels, $\lambda_1=5$, and $\lambda_2=10$. $\Phi(\Lambda^{\textrm{new}}_{v})$ denotes the fast-RCNN's detection score for the patch of $\Lambda^{\textrm{new}}_{v}$.}. Our method learned AOGs with about $1/6$--$1/2$ part annotations, but exhibited superior performance to the second best baseline.

\subsection{Justification of the methodology}

We have three reasons to explain the good performance of our method. First, \textbf{generic information}: the latent patterns in the AOG were pre-fine-tuned using massive object images in a category, instead of being learned from a few part annotations. Thus, these patterns reflected generic part appearances and did not over-fit to a few part annotations.

Second, \textbf{less model drifts:} Instead of learning new CNN parameters, our method just used limited part annotations to mine the related patterns to represent the part concept. In addition, during active QA, Equation~(\ref{eqn:predict}) usually selected objects with common poses for QA, \emph{i.e.} choosing objects sharing common latent patterns with many other objects. Thus, the learned AOG suffered less from the model-drift problem.

\begin{table}[t]
\begin{center}
\caption{Part localization evaluated using the PCP metric.}
\label{tab:pcp}
\resizebox{\linewidth}{!}{\begin{tabular}{lccc}
\hline
\!\!\!&\!\!\!\!\!\!\!\!\!\!\!\! {\small\# of part annot.} \!\!\!\!&\!\!\!\! {\small VOC Part} \!\!\!\!&\!\!\!\! {\small ILSVRC Animal}\!\!\!\!\\
\!\!\!{\small SS-DPM-Part~\cite{SSDPM}} & 60 & 7.2 & 19.2\\
\!\!\!{\small PL-DPM-Part~\cite{PLDPM}} & 60 & 6.7 & 12.8\\
\!\!\!{\small Part-Graph~\cite{SemanticPart}} & 60 & 11.0 & 25.6\\
\!\!\!{\small fc7+linearSVM} & 60 & 13.5 & 24.5\\
\!\!\!{\small fc7+RBF-SVM} & 60 & 9.5 & 18.8\\
\!\!\!{\small VAE+linearSVM~\cite{VAE}} & 30 & 6.7 & --\\
\!\!\!{\small CoopNet+linearSVM~\cite{CoopNet}} & 30 & 5.6 & --\\
\!\!\!{\small Fast-RCNN (1 ft)~\cite{FastRCNN}} & 30 & 34.5 & 62.3\\
\!\!\!{\small Fast-RCNN (2 fts)~\cite{FastRCNN}} & 30 & 45.7 & 68.6\\
\!\!\!{\small Ours+fastRCNN} & \textcolor{red}{\bf 10} & 33.0 & 53.0\\
\!\!\!{\small Ours+fastRCNN} & \textcolor{red}{\bf 20} & {\bf47.2} & 64.9\\
\!\!\!{\small Ours+fastRCNN} & \textcolor{red}{\bf 30} & {\bf50.5} & {\bf 71.1}\\
\hline
\end{tabular}}
\end{center}
\end{table}

Third, \textbf{high QA efficiency:} Our QA process balanced both the commonness and the accuracy of a part template in Equation~(\ref{eqn:predict}). In early steps of QA, our approach was prone to asking about new part templates, because objects with un-modeled part appearance usually had low inference scores. In later QA steps, common part appearances had been modeled, and our method gradually changed to ask about objects belonging to existing part templates to refine the AOG. Our method did not waste much labor of labeling objects that had been well modeled or had strange appearance.

\section{Summary and discussion}

In this paper, we have proposed a method to bridge and solve the following three crucial issues in computer vision simultaneously.
\begin{itemize}
\item Removing noisy representations in conv-layers of a CNN and using an AOG model to reveal the semantic hierarchy of objects hidden in the CNN.
\item Enabling people to communicate with neural representations in intermediate conv-layers of a CNN directly for model learning, based on the semantic representation of the AOG.
\item Weakly-supervised transferring of object-part representations from a pre-trained CNN to model object parts at the semantic level, which boosts the learning efficiency.
\end{itemize}

Our method incrementally mines object-part patterns from conv-layers of a pre-trained CNN and uses an AOG to encode the mined semantic hierarchy. The AOG semanticizes neural units in intermediate feature maps of a CNN by associating these units with semantic parts. We have proposed an active QA strategy to learn such an AOG model in a weakly-supervised manner. We have tested the proposed method for a total of 37 categories in three benchmark datasets. Our method has outperformed other baselines in the application of weakly-supervised part localization. For example, our method with 11 part annotations performed better than fast-RCNN with 60 part annotations on the ILSVRC dataset in Fig.~\ref{fig:curve}.

\ifCLASSOPTIONcompsoc
  \section*{Acknowledgments}
\else
  \section*{Acknowledgment}
\fi
This work is supported by ONR MURI project N00014-16-1-2007, DARPA XAI Award N66001-17-2-4029, and NSF IIS 1423305.

\ifCLASSOPTIONcaptionsoff
  \newpage
\fi

\bibliographystyle{IEEE}
\bibliography{TheBib}

\vspace{-30pt}
\begin{IEEEbiography}[{\includegraphics[width=1in,height=1.25in,clip,keepaspectratio]{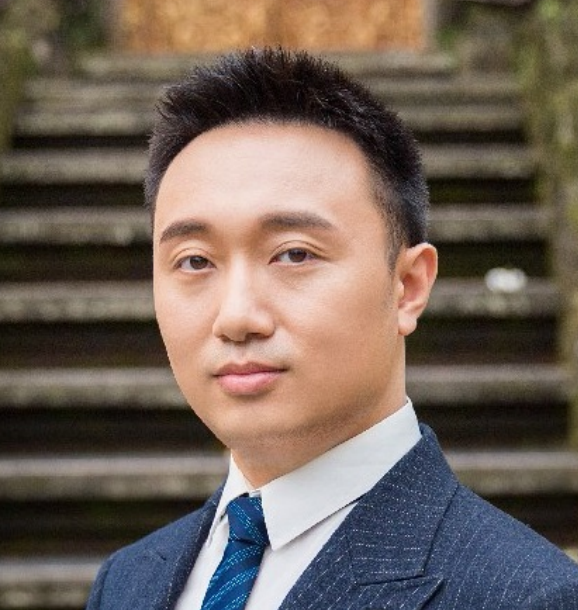}}]{Quanshi Zhang}
received the B.S. degree in machine intelligence from Peking University, China, in 2009 and M.S. and Ph.D. degrees in center for spatial information science from the University of Tokyo, Japan, in 2011 and 2014, respectively. In 2014, he went to the University of California, Los Angeles, as a post-doctoral associate. Now, he is an associate professor at the Shanghai Jiao Tong University. His research interests include computer vision, machine learning, and robotics.
\end{IEEEbiography}

\vspace{-30pt}
\begin{IEEEbiography}[{\includegraphics[width=1in,height=1.25in,clip,keepaspectratio]{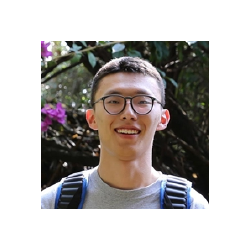}}]{Ruiming Cao}
received the B.S. degree in computer science from the University of California, Los Angeles, in 2017. Now, he is a master student at the University of California, Los Angeles. His research mainly focuses on computer vision.
\end{IEEEbiography}

\vspace{-30pt}
\begin{IEEEbiography}[{\includegraphics[width=1in,height=1.25in,clip,keepaspectratio]{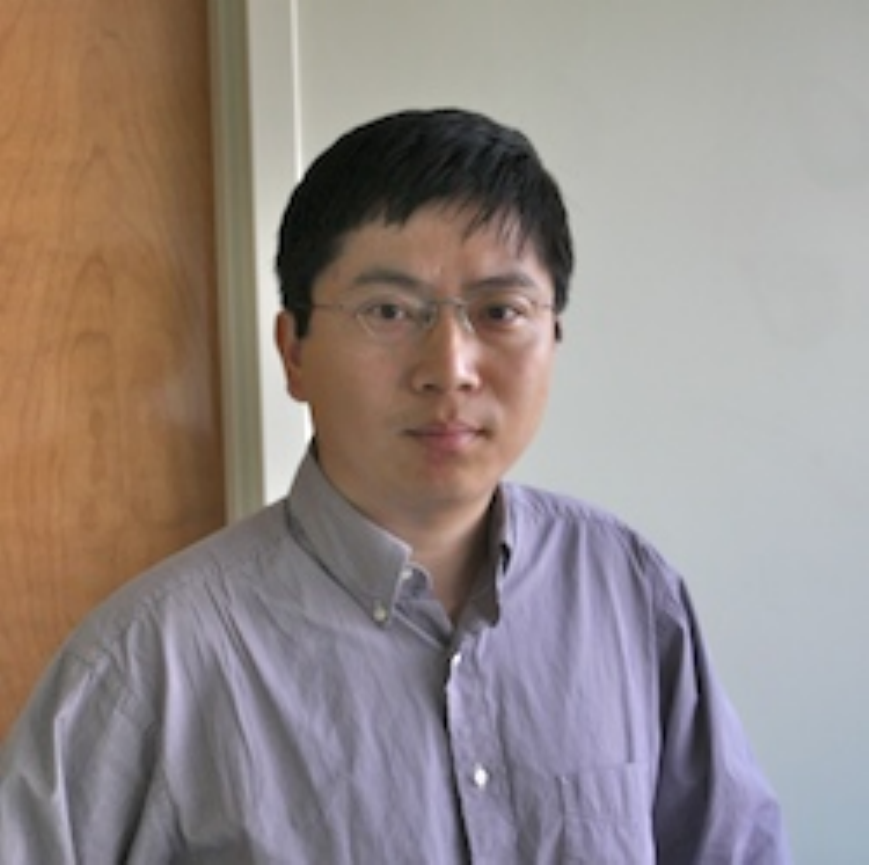}}]{Ying Nian Wu}
received a Ph.D. degree from the Harvard University in 1996. He was an Assistant Professor at the University of Michigan between 1997 and 1999 and an Assistant Professor at the University of California, Los Angeles between 1999 and 2001. He became an Associate Professor at the University of California, Los Angeles in 2001. From 2006 to now, he is a professor at the University of California, Los Angeles. His research interests include statistics, machine learning, and computer vision.
\end{IEEEbiography}

\vspace{-30pt}
\begin{IEEEbiography}[{\includegraphics[width=1in,height=1.25in,clip,keepaspectratio]{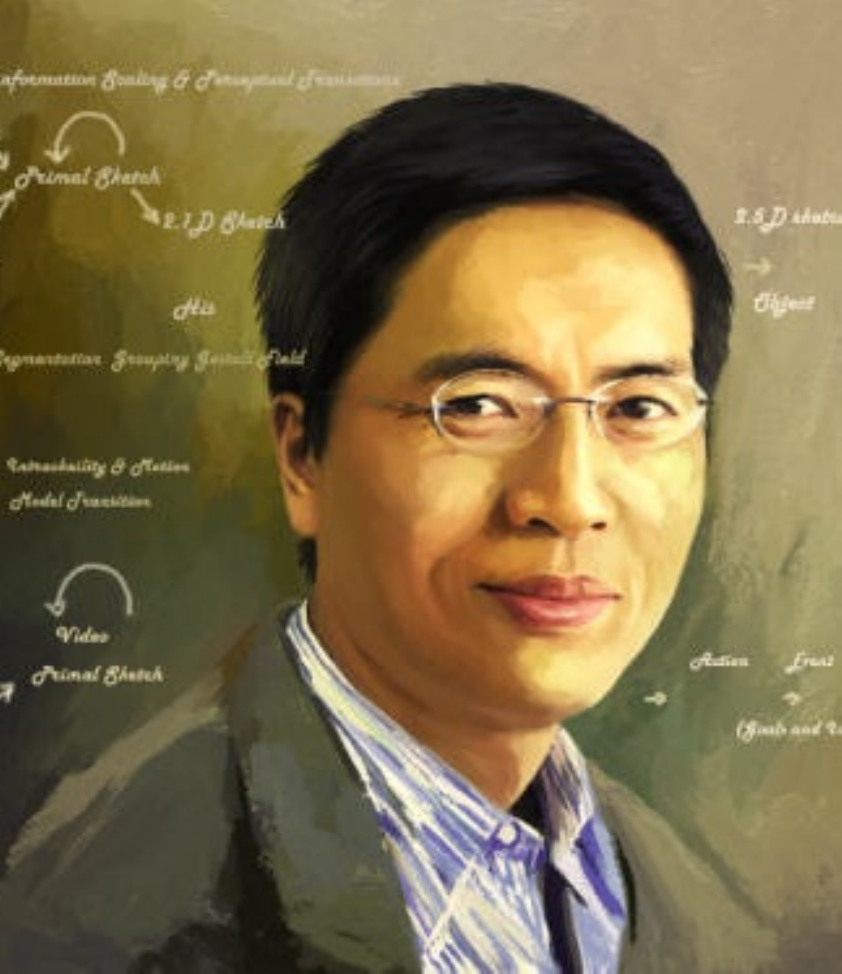}}]{Song-Chun Zhu}
Song-Chun Zhu received a Ph.D. degree from Harvard University, and is a professor with the Department of Statistics and the Department of Computer Science at UCLA. His research interests include computer vision, statistical modeling and learning, cognition and AI, and visual arts. He received a number of honors, including the Marr Prize in 2003 with Z. Tu et. al. on image parsing,the Aggarwal prize from the Int'l Association of Pattern Recognition in 2008, twice Marr Prize honorary nominations in 1999 for texture modeling and 2007 for object modeling with Y.N. Wu et al., a Sloan Fellowship in 2001, the US NSF Career Award in 2001, and the US ONR Young Investigator Award in 2001. He is a Fellow of IEEE.
\end{IEEEbiography}
\vfill

\newpage
\onecolumn
\appendices
\section*{And-Or graph representations}

\subsection*{Parameters for latent patterns}

We use the notation of ${\bf p}_{u}$ to denote the central position of an image region $\Lambda_{u}$. For simplification, all position variables ${\bf p}_{u}$ are measured based on the image coordinates by propagating the position of $\Lambda_{u}$ to the image plane.

Each latent pattern $u$ is defined by its location parameters $\{L_{u},D_{u},\overline{\bf p}_{u},\Delta{\bf p}_{u}\}\subset{\boldsymbol\theta}$, where ${\boldsymbol\theta}$ is the set of AOG parameters. It means that a latent pattern $u$ uses a square within the $D_{u}$-th channel of the $L_{u}$-th conv-layer's feature map as its deformation range. The center position of the square is given as $\overline{\bf p}_{u}$. When latent pattern $u$ is extracted from the $k$-th conv-layer, $u$ has a fixed value of $L_{u}=k$.

$\Delta{\bf p}_{u}$ denotes the average displacement from $u$ and $u$'s parent part template $v$ among various images, and $\Delta{\bf p}_{u}$ is used to compute $S^{\textrm{inf}}(\Lambda_{u}|\Lambda_{v})$. Given parameter $\overline{\bf p}_{u}$, the displacement $\Delta{\bf p}_{u}$ can be estimated as
\begin{equation}
\Delta{\bf p}_{u}=\overline{\bf p}^{*}_{v}-\overline{\bf p}_{u}\nonumber
\end{equation}
where $\overline{\bf p}^{*}_{v}$ denotes the average position of all ground-truth parts that are annotated for part template $v$. As a result, for each latent pattern $u$, we only need to learn its channel $D_{u}\in{\boldsymbol\theta}$ and central position $\overline{\bf p}_{u}\in{\boldsymbol\theta}$.

\subsection*{Scores of terminal nodes}

The inference score for each terminal node $v^{\textrm{unt}}$ under a latent pattern $u$ is formulated as
\begin{small}
\begin{eqnarray}
&S_{v^{\textrm{unt}}}=S_{v^{\textrm{unt}}}^{\textrm{rsp}}+S_{v^{\textrm{unt}}}^{\textrm{loc}}+S_{v^{\textrm{unt}}}^{\textrm{pair}}\nonumber\\
&S_{v^{\textrm{unt}}}^{\textrm{rsp}}=\left\{\begin{array}{ll}\lambda^{\textrm{rsp}}X(v^{\textrm{unt}}),& X(v^{\textrm{unt}})>0\\ \lambda^{\textrm{rsp}}S_{none},& X(v^{\textrm{unt}})\leq0\end{array}\right.\nonumber\\
&S_{v^{\textrm{unt}}}^{\textrm{pair}}=-\lambda^{\textrm{pair}}\!\!\!\!\!\!\!\!\underset{u_{\textrm{upper}}\in\!\textrm{Neighbor}(u)}{\mathbb{E}}\!\!\!\!\!\!\Vert[{\bf p}_{v^{\textrm{unt}}}-{\bf p}_{u_{\textrm{upper}}}]-[\overline{\bf p}_{u_{\textrm{upper}}}-\overline{\bf p}_{u}]\Vert\nonumber
\end{eqnarray}
\end{small}
The score of $S_{v^{\textrm{unt}}}$ consists of the following three terms: 1) $S_{v^{\textrm{unt}}}^{\textrm{rsp}}$ denotes the response value of the unit $v^{\textrm{unt}}$, when we input image $I$ into the CNN. $X(v^{\textrm{unt}})$ denotes the normalized response value of $v^{\textrm{unt}}$; $S_{none}=-3$ is set for non-activated units. 2) When the parent $u$ selects $v^{\textrm{unt}}$ as its location inference (\emph{i.e.} $\Lambda_{u}\leftarrow\Lambda_{v^{\textrm{unt}}}$), $S_{v^{\textrm{unt}}}^{\textrm{loc}}$ measures the deformation level between $v^{\textrm{unt}}$'s location ${\bf p}_{v^{\textrm{unt}}}$ and $u$'s ideal location $\overline{\bf p}_{u}$. 3) $S_{v^{\textrm{unt}}}^{\textrm{pair}}$ indicates the spatial compatibility between neighboring latent patterns: we model the pairwise spatial relationship between latent patterns in the upper conv-layer and those in the current conv-layer. For each $v^{\textrm{unt}}$ (with its parent $u$) in conv-layer $L_{u}$, we select 15 nearest latent patterns in conv-layer $L_{u}+1$, \emph{w.r.t.} $\Vert\overline{\bf p}_{u}-\overline{\bf p}_{u_{\textrm{upper}}}\Vert$, as the neighboring latent patterns. We set constant weights $\lambda^{\textrm{rsp}}=1.5,\lambda^{\textrm{loc}}=1/3,\lambda^{\textrm{pair}}=10.0$, $\lambda^{\textrm{unant}}=5.0$, and $\lambda^{\textrm{close}}=0.4$ for all categories. Based on the above design, we first infer latent patterns corresponding to high conv-layers, and use the inference results to select units in low conv-layers.

During the learning of AOGs, we define $S^{\textrm{unant}}_{u}=S_{\hat{v}^{\textrm{unt}}}^{\textrm{rsp}}+S_{\hat{v}^{\textrm{unt}}}^{\textrm{loc}}$ to measure the latent-pattern-level inference score in Equation~(5), where $\hat{v}^{\textrm{unt}}$ denotes the neural unit assigned to $u$.

\subsection*{Scores of AND nodes}

\begin{equation}
S^{\textrm{inf}}(\Lambda_{u}|\Lambda_{v})=-\lambda^{\textrm{inf}}\min\{\Vert{\bf p}(\Lambda_{u})+\Delta{\bf p}_{u}-{\bf p}(\Lambda_{v})\Vert^2,d^2\}\nonumber
\end{equation}
where we set $d=37$ pixels and $\lambda^{\textrm{inf}}=5.0$.

\end{document}